\pdfoutput=1

\documentclass[11pt]{article}

\usepackage[preprint]{acl}

\usepackage{times}
\usepackage{latexsym}
\usepackage{inconsolata}

\usepackage[utf8]{inputenc} 
\usepackage[T1]{fontenc}    
\usepackage{xtab, booktabs}       
\usepackage{amsfonts}       
\usepackage{nicefrac}       
\usepackage{xcolor}         
\usepackage{natbib}
\usepackage{xspace}
\usepackage{dsfont}
\usepackage{amsmath,amssymb}
\usepackage[ruled,norelsize]{algorithm2e}
\usepackage{multirow}
\usepackage{booktabs}
\usepackage{graphicx}
\usepackage{subcaption}
\usepackage{wrapfig}
\usepackage{enumerate}
\usepackage{enumitem}
\usepackage{url}
\usepackage{setspace}
\usepackage{arydshln}
\usepackage{hyperref}
\usepackage{float}
\usepackage{longtable}

\usepackage{amsmath,amsfonts,bm}

\newcommand{\anB}[1]{\langle #1 \rangle}

\def\vx{{\bm{x}}}

\DeclareMathAlphabet{\mathsfit}{\encodingdefault}{\sfdefault}{m}{sl}
\SetMathAlphabet{\mathsfit}{bold}{\encodingdefault}{\sfdefault}{bx}{n}

\def\cA{{\mathcal{A}}}

\def\cE{{\mathcal{E}}}

\def\cT{{\mathcal{T}}}

\newif\ifcomments
\commentstrue

\newcommand{\ttt}[1]{\texttt{#1}}

\newcommand{\method}{{\sc Tiger-8B}\xspace}
\newcommand{\methodhead}{{\sc Tiger}\xspace}
\newcommand{\dataset}{\ttt{ReWild}\xspace}
\newcommand{\llm}{f_\text{LLM}}

\title{Can LLMs Reason in the Wild with Programs?}

\author{Yuan Yang\textsuperscript{1}, Siheng Xiong\textsuperscript{1}, Ali Payani\textsuperscript{2}, Ehsan Shareghi\textsuperscript{3} \& Faramarz Fekri\textsuperscript{1}\\
\textsuperscript{1}Georgia Institute of Technology, 
\textsuperscript{2}Cisco,
\textsuperscript{3}Monash University\\
\ttt{\{yyang754@,sxiong45@,faramarz.fekri@ece.\}gatech.edu} \\
\ttt{apayani@cisco.com}\ \ \ 
\ttt{ehsan.shareghi}@monash.edu
}

\begin{document}
\maketitle
\begin{abstract}

Large Language Models (LLMs) have shown superior capability to solve reasoning problems with programs. While being a promising direction, most of such frameworks are trained and evaluated in settings with a prior knowledge of task requirements. However, as LLMs become more capable, it is necessary to assess their reasoning abilities in more realistic scenarios  
%
%
where many real-world problems are open-ended with ambiguous scope,  and often require multiple formalisms to solve. 
%
To investigate this, we introduce the \textit{reasoning in the wild} task, where an LLM is tasked to solve a reasoning problem of unknown type by identifying the sub-problems and their corresponding formalisms, and writing a program to solve each sub-problem, guided by a \textit{tactic}.
%
We create a large tactic-guided trajectory dataset containing detailed solutions to a diverse set of reasoning problems, ranging from well-defined single-form reasoning (e.g., math, logic), to ambiguous and hybrid ones (e.g., commonsense, combined math and logic). This allows us to test various aspects of LLMs reasoning at the fine-grained level such as the selection and execution of tactics, and the tendency to take undesired shortcuts.
In experiments, we highlight that existing LLMs fail significantly on problems with ambiguous and mixed scope, revealing critical limitations and overfitting issues (e.g. accuracy on GSM8K drops by at least 50\%). We further show fine-tuning a local LLM on the trajectories data leads to better performance.
Project repo is available \href{https://github.com/gblackout/Reason-in-the-Wild}{here}.



\end{abstract}

\section{Introduction}

\begin{figure}[t]
    \centering
    \includegraphics[trim={1cm 0cm 5cm 1cm},clip,width=\linewidth]{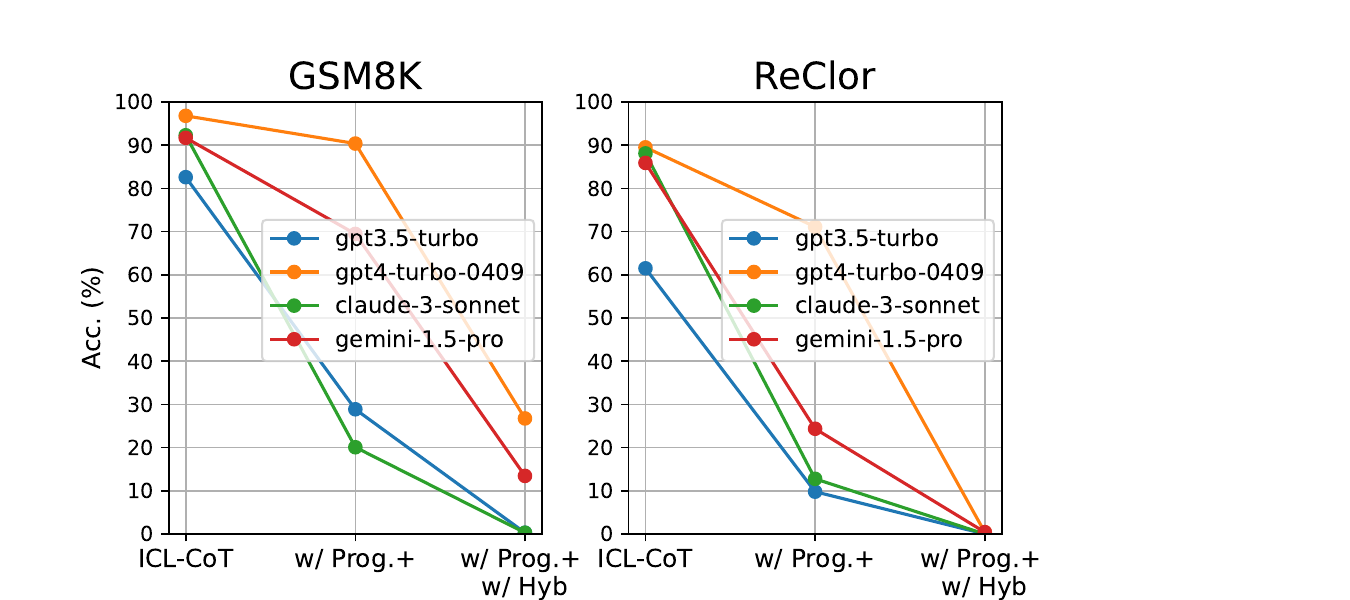}
    \caption{
    Commonly used metrics such as accuracy, while suggesting LLMs perform well on reasoning tasks in free-form (\textit{ICL-CoT}), fail to reflect their capability in a more fine-grained reasoning benchmark. 
    We find LLMs' accuracy drops significantly when tasked to solve problems with a non-trivial program (\textit{w/ Prog.+}); this further decreases when tasked to reason in the wild, where the tasks are blended with other contexts and their reasoning types are unknown (\textit{w/ Hyb}). }
    \label{fig:showoff-fig}
    \vspace{-.2cm}
\end{figure}


Large language models (LLMs) demonstrate strong capabilities in solving a wide variety of reasoning problems that involve different formalisms such as formal logic, math, graph, and commonsense reasoning.
As an example, recent LLMs achieve above 90\% (Figure~\ref{fig:showoff-fig}) accuracy on math benchmarks such as GSM8K~\citep{cobbe2021training}, and above 80\% accuracy on commonsense reasoning benchmarks such as ReClor~\citep{yu2020reclor}.
However, these results are typically obtained in the in-context learning (ICL) setting with answers generated in free text.
Such a way is criticized for lacking reliability and is prone to question distribution shift and hallucination~\citep{kabra_program-aided_2023, schick_toolformer_2023,hao_toolkengpt_2023}.

\begin{figure*}[t]
    \centering
    \includegraphics[trim={0cm 1.9cm .5cm 0cm},clip,width=\textwidth]{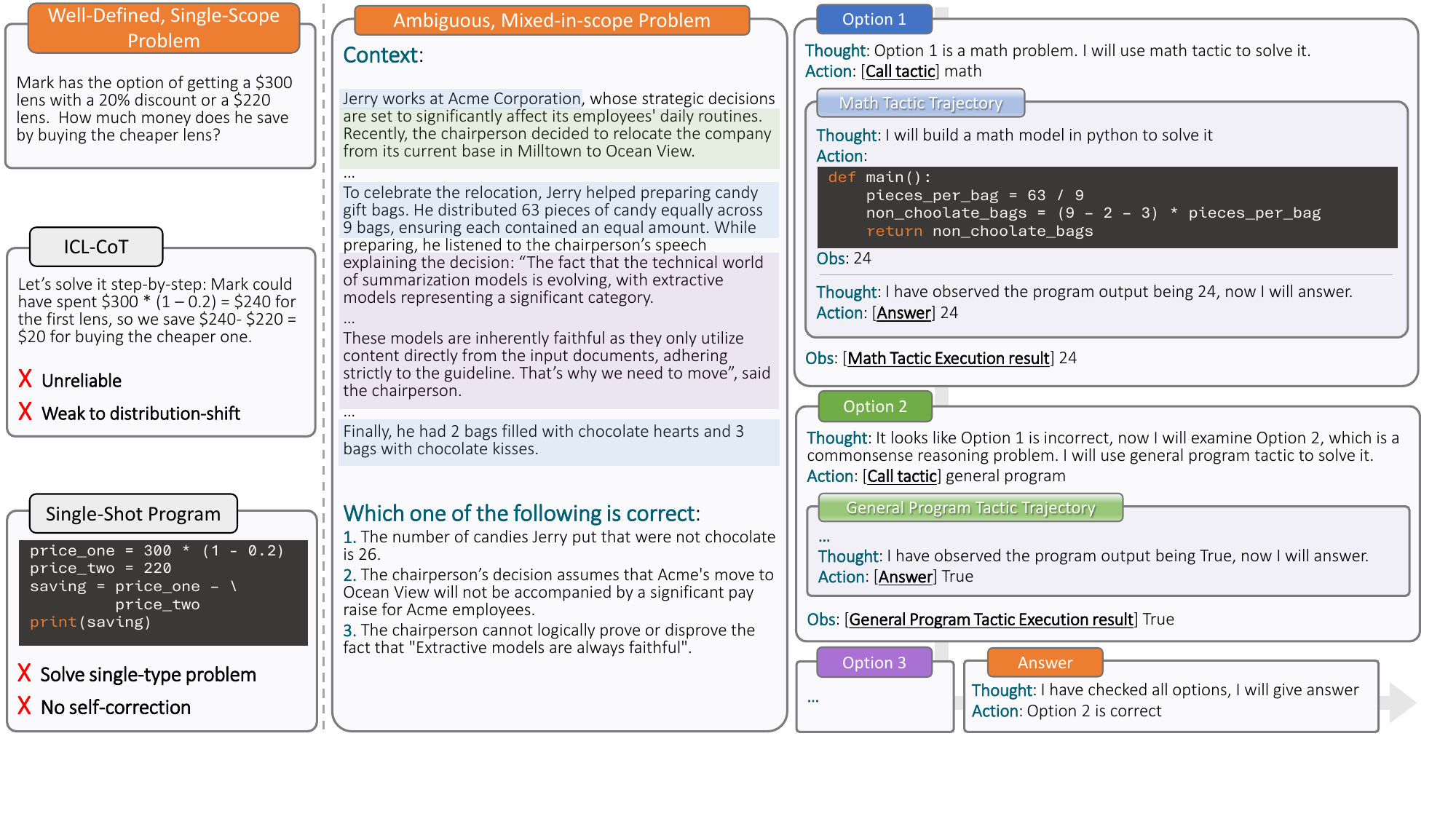}
    \caption{Solving ambiguous, mixed-in-scope problem via reasoning in the wild.}
    \label{fig:framework-overview-fig}
    \vspace{-.1cm}
\end{figure*}

Agent-like\footnote{We use "Agent-like" as many of the existing approaches that augment LLMs with external mechanisms only have partial components of a language agent~\cite{coala}.} LLM frameworks show great potential in addressing these limitations, which solves reasoning tasks by writing programs~\citep{yao2022react,gou2023tora,yuan_advancing_2024}.
While being a promising direction, most of these frameworks are trained and evaluated on benchmarks with well-defined scope and a clear formalism to solve, such as GSM8K, where questions are guaranteed to be an arithmetic problem that can be solved with a single math model.
Such an assumption does not hold in real-world scenarios, where the problems could:
(1) \textbf{be ambiguous in scope}, where the problem is not a well-defined math or formal logic task, and the program written cannot fully solve the problem and requires further reasoning to derive the final answer;
(2) \textbf{require multiple formalisms to solve}, where the problem needs to be decomposed into subproblems with each requiring different sets of skills to solve;
(3) \textbf{be mixed with irrelevant context}, where the context contains irrelevant information that needs to be excluded when building a formal model. Such problems are widely present in the real world, for example, Figure~\ref{fig:framework-overview-fig} shows an example of a multichoice question, where
Option 1 requires a math model to verify, 
Option 2 requires a generic program, which does not fit into any of the well-defined formalisms, 
and Option 3 requires a predicate logic model to verify.
Solving this problem poses a higher demand on the \textit{reasoning-in-the-wild} capability of an LLM agent, who needs to recognize the subproblems, identify the formalism, and write the program to finally answer it.
Yet, there lacks such a comprehensive benchmark to evaluate LLMs' capability in these aspects.

To move towards bridging this gap, we introduce the task of \emph{reasoning in the wild}.
Given a reasoning problem of an \emph{unknown type}, the 
task is to solve the problem by iteratively identifying the sub-problems and their corresponding formalisms, then writing the suitable programs to solve each subproblem. To better control and assess the agent's behavior, we introduce the notion of~\textit{tactics}. We draw inspiration from the tactic concept in interactive theorem prover literature, where a person proves a math theorem by decomposing and transforming the problem into sub-problems and solving them by corresponding tactics~\citep{lean4, isabelle}.
In our setting, a tactic is a guideline consisting of descriptions, code templates, and an action space, which defines a specific way to tackle the given problem. As we will see in later sections, by explicitly applying the constraint of tactic we obtain a fine-grained insight into LLMs' behavior in problem-solving beyond common holistic evaluation protocols. 



To enable such a study, we create a large tactic-guided trajectory dataset, namely~\dataset, that consists of problem-solving trajectories (formatted similar to ReAct~\citep{yao2022react} as a chain of \ttt{Thought}, \ttt{Action}, and corresponding \ttt{Observation}) over a wide range of reasoning problems. These problems range from well-defined single-form reasoning such as GSM8K~\citep{cobbe2021training}, FOLIO~\citep{han2022folio} to ambiguous one such as ReClor~\citep{yu2020reclor}.
On top of existing problems, we also create new hybrid problems that require multiple formalisms to solve.
\dataset is generated through GPT4 with expert-written prompts and is further post-processed and verified by a combined pipeline of manual annotation and automated filtering.
The resulting dataset consists of 6.7K trajectories with a total of 21.7M tokens.

In our experiments, we evaluate a diverse set of the most powerful LLMs to date on our benchmark.
Remarkably, we find existing LLMs fail significantly when tasked to solve reasoning problems with tactic-guided programs (see Figure~\ref{fig:showoff-fig}, \textit{Prog.+}) and the performance further deteriorates for hybrid problems (see Figure~\ref{fig:showoff-fig}, \textit{Hyb}).
Through the lens of tactics, \textbf{we analyze the results and identify three critical limitations of existing LLMs}:
(1) Many LLMs show ``overfitted'' behavior and fail to follow the tactic on popular problems such as GSM8K, leading to a drop in performance;
(2) Most LLMs, except for GPT4 series, show a lack of the capability of ``instruction-following in long context'', where it fails to follow the tactic on trajectories that are typically 3K long;
(3) Powerful LLMs including GPT4 tend to hallucinate and generate ``trivial programs'' on ambiguous reasoning problems, showing a poor generalizability over out-of-distribution problems.
Finally, we show that these limitations can be alleviated via fine-tuning. We train and release a LLaMA3-8B model on \dataset, which we refer to as \textbf{T}act\textbf{I}c-\textbf{G}uided Reason\textbf{ER} (\method), and show it achieves GPT4 level performance.

\section{Related Work}




\textbf{Program-aided LLMs}.
Recent research has improved LLMs' reasoning capabilities with the help of programs, achieving better performance on 
math~\citep{gao_pal_2023,kabra_program-aided_2023, chen_program_2023}
and logic~\citep{feng_language_2023, pan_logic-lm_2023, yang2023harnessing, olausson_linc_2023, ye_satlm_nodate}
reasoning tasks.
These works are ``hardwired'' to solve a specific type of reasoning problem, and do not explicitly model the interactions with the environment as a trajectory, limiting their in-the-wild applications.





\textbf{Trajectory-based LLM agents}
Agent-like LLMs models such as ReAct~\citep{yao2022react} explicitly model the interaction as a trajectory, and several works study its potential in solving reasoning problems.
The most prominent ones in this space are FireAct~\citep{chen2023fireact}, ToRA~\citep{gou2023tora},
and EURUS~\citep{yuan_advancing_2024},
which collect problem-solving trajectory data from GPT models in solving various known reasoning tasks and train local models, showing they can achieve better performance after fine-tuning. 

These approaches highlight the benefits of programs or fine-tuning but do not uncover critical limitations of existing LLMs. Our work underscores several key aspects neglected in previous works. We demonstrate: (1) how to set up a unified reasoning framework to tackle various existing and new (mixed) reasoning problems, (2) how to incorporate various mechanisms (\S\ref{sec:wild}) in the reasoning process which will enable a more fine-grained analysis of system's abilities, and (3) how to evaluate such complex trajectories to gain deeper insight into LLMs' limitations and behavior beyond common holistic evaluation protocols. 

\section{Reasoning in the Wild}\label{sec:wild}

\begin{figure}[t]
    \centering
    \includegraphics[trim={0cm 1.8cm 15cm 0cm},clip,width=\linewidth]{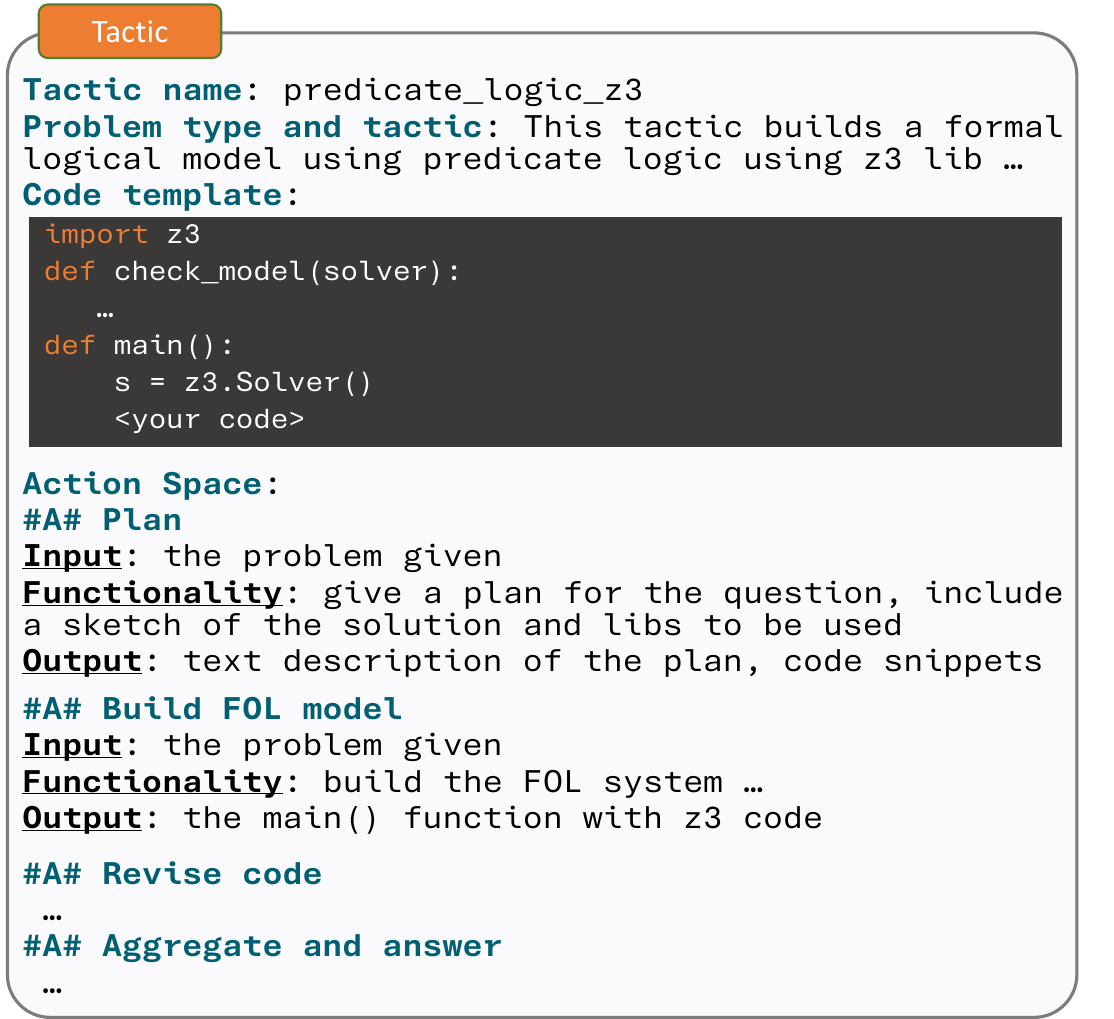}
    \caption{Example of a predicate logic tactic.}
    \label{fig:tactic-doc}
    \vspace{-.1cm}
\end{figure}

The task of \textit{reasoning in the wild} considers an LLM agent solving reasoning problems by writing programs and interacting with an environment defined by a \textit{tactic}.
Similar to prior agent-like LLM frameworks~\citep {yao2022react}, the problem-solving process consists of a series of \ttt{Thoughts}, \ttt{Actions}, and \ttt{observations}.
At each step, the LLM agent $\llm$ generates a thought $t$ that serves as a chain-of-thought (CoT) context for its action, and an action $a$ which carries out the actual action in the environment: this could be outputting a plan in free text (i.e., \ttt{Plan}), generating a program (i.e., \ttt{Write Program}), or returning an answer (i.e., \ttt{Answer}).
Then it receives observation $o$ from the environment.
The series of $\cE = [\anB{t,a,o}^{(1)}, \anB{t,a,o}^{(2)}, ...]$ forms a problem-solving \textit{trajectory} $\cE$ that will be included as prompt for the next round of interaction.

Different from the prior work, \textbf{we extend this problem-solving trajectory by explicitly applying a tactic to specify and monitor the agent's behavior}.
Formally, let
$T \in \cT = \{T_1, T_2, ...\}$
be a tactic, and each tactic is defined by
$T = \{\vx_T, \cA, O\}$ (Figure~\ref{fig:tactic-doc}). Here,
$\vx_T$ is the tactic description, which is a passage describing what this tactic is suitable for and how it is generally used;
$\cA = \{a_1, a_2, ...\}$ is the action space defining the allowable actions in this tactic, for example, \ttt{Plan} and \ttt{Write Program}. Each action is defined by its input, functionality, and output $a = \{ a_\text{in}, a_\text{func}, a_\text{out}\}$;
and $O$ is the tactic-corresponding observer that parses, monitors, executes the agent's action, and finally provides observations $o$. 
A complete tactic is shown in \S\ref{app:method-examples}.


\begin{figure}[t]
    \centering
    \includegraphics[trim={0cm 6.2cm 20.1cm 0cm},clip,width=.8\linewidth]{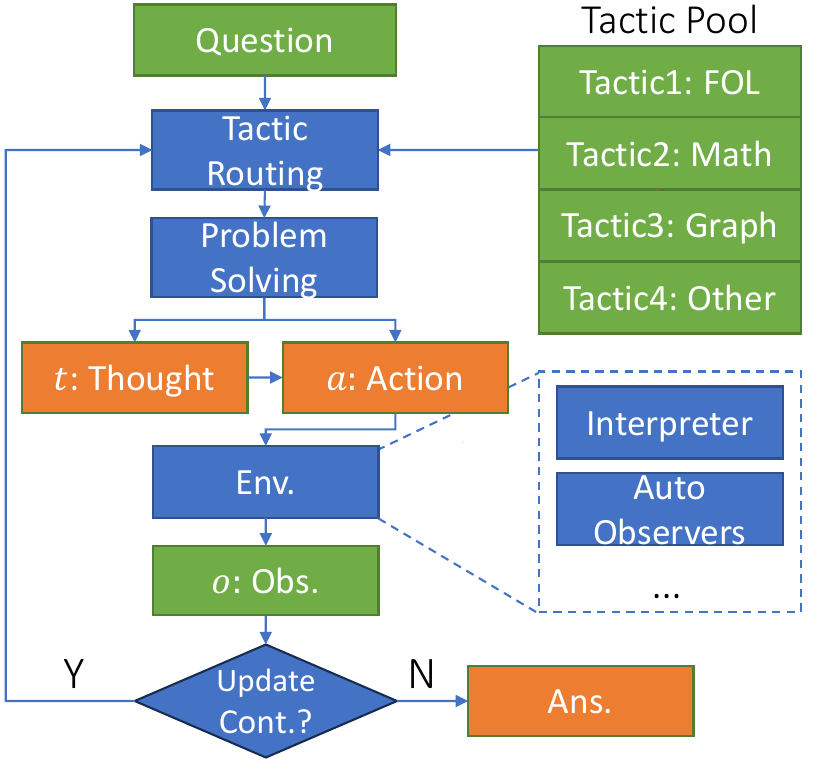}
    \caption{Solving reasoning problems by applying \ttt{Routing} and problem specific tactics.}
    \label{fig:tactic-flow-chart}
\end{figure}
\vspace{-.1cm}

Let $q$ be a reasoning problem and $\llm$
be the LLM agent, the agent solves the problem by performing two levels of reasoning:
(1) At the top level, a \ttt{Routing} trajectory is formed that picks the subproblems and the corresponding tactics;
(2) At the second level, the chosen tactic is carried out, forming a ``sub-trajectory'' for subproblem solving. As shown in Figure~\ref{fig:tactic-flow-chart}:

\noindent\textbf{1. Routing}: Given the problem $q$, the agent executes the \ttt{Routing} tactic, that is $T_r$, where it extracts the subproblem $q_\text{sub}$ from $q$, and identifies the best tactic $T$ to solve it by checking all the tactic descriptions included in $T_r$ and reflecting on the past experience $\cE$ (where $\cE^{(0)} = \emptyset$). At every $(i)$-th iteration:
\begin{equation*}
    t_r^{(i)}, a_r^{(i)} \gets \llm(q, T_r, \cE^{(i-1)}),
\end{equation*}
and $[q_\text{sub}, T_\text{sub}]$ are parsed from the output part of the action $a_r^{(i)}$ (i.e., $a_{r, \text{out}}^{(i)}$).

\noindent\textbf{2. Problem solving}: Solves subproblem $q_\text{sub}$ with the chosen tactic $T_\text{sub}$. At every $(j)$-th iteration:
\begin{align*}
    & t^{(j)}, a^{(j)} \gets \llm(q_\text{sub}, T_\text{sub}, \cE_\text{sub}^{(j-1)}), \\
    & o^{(j)} \gets O(a^{(j)}),\
    \cE_\text{sub}^{(j)} = \cE_\text{sub}^{(j-1)} \cup \{\anB{t, a, o}^{(j)}\},
\end{align*}
where sub-trajectory $\cE_\text{sub}^{(0)} = \emptyset$. The sub-trajectory terminates if the \ttt{Answer} action is called where the output part $a_\text{out}$ is returned as the observation $o_r^{(i)}$ to the \ttt{Routing} trajectory.

\noindent\textbf{3. Update Routing}: Update the routing trajectory
\begin{equation*}
    \cE^{(i)} = \cE^{(i-1)} \cup \{\anB{t, a, o}_r^{(i)}\},
\end{equation*}
and then repeat step 1 until the \ttt{Answer} action is called in the routing trajectory, where the final answer to problem $q$ is aggregated and returned.

By explicitly applying and monitoring the tactics and actions, we
(1) make it easy to control the agent's behavior, defending against potentially undesired behaviors;
(2) gain deeper insight into the LLMs, by tracing if it uses the right tactic for the right subproblems;
(3) make it possible to create a challenging hybrid problem that requires multiple tactics to solve with fine-grained metrics.










\section{Data Generation}
\label{sec:data-gen}
\vspace{-.1cm}

To enable evaluating (and training) LLMs on our task,
a large dataset of problem-solving trajectories over diverse reasoning problems needs to be created.
To do so, we select several existing datasets and generate the trajectories with the specific tactic and ICL prompts and select those successful ones with \emph{non-trivial programs}~(introduced shortly) to include in our dataset, \dataset.

\textbf{Tactics and datasets}.
We manually create a pool of diverse tactics, each corresponding to the following datasets:
(1) \textbf{Math Tactic} for GSM8K;
(2) \textbf{Logic Tactic} for FOLIO, which is a logic-grounded natural language inference (NLI) dataset;
(3) \textbf{Graph Tactic} for ProScript~\citep{sakaguchi2021proscript}, which is a dataset containing graph-like steps for achieving certain goals such as ``opening a bank account'';
and (4) \textbf{General Program Tactic} for ReClor, which is a commonsense reasoning dataset.
To generate trajectories, we manually created examples and included them as ICL prompts to guide the generation using a mixture of three models:
\ttt{gpt4-turbo-0409}, \ttt{gpt-4o}, and \ttt{claude-3-opus}. More details in~\S\ref{app:data-gen-examples}.

\textbf{Post-processing}.
We filter the generated trajectories with the following steps:
(1) We filter those that do not write any programs or those with programs that failed to run.
This indicates the model ``shortcuts'' the process by directly outputting the answer.
(2) A more severe shortcutting happens with ReClor dataset for all the LLMs we tested.
We refer to this as ``\textbf{trivial programs}'':
ReClor problems are typically ambiguous in scope and do not fit into any existing formalisms are known to be trained by these LLMs.
When LLMs are ``forced'' to generate programs, they often generate programs that ``hardcoded'' the answer and put CoT free-form reasoning in the comments.
We filter them by labeling a set of programs and using them as ICL prompts of an LLM classifier (Examples and details are provided in \S\ref{app:data-gen-examples}).
In experiments, we show this phenomenon is widely present and causes a significant performance drop.



\textbf{Hybrid data generation}.
Apart from the problem-solving trajectories, we also generate hybrid problems and routing trajectories.
As an example, let
$q_1 = [q_{c,1}, q_{a,1}]$
and $q_2 = [q_{c,2}, q_{a,2}]$
be two problems randomly sampled from the original datasets, where $q_c$ is the ``context'' part and $q_a$ the ``answer'' part of the question.
We create a multichoice hybrid problem by
(1) ``blending'' the context part into a new coherent passage using an LLM (i.e., \ttt{gpt4-turbo-0409})
$q^*_{c} = \llm(q_{c,1}, q_{c,2})$;
and (2) putting each answer $q_{a,i}$ as an \textit{option} to the new hybrid problem with only one of them correct and the rest replaced with incorrect answers.
Figure~\ref{fig:framework-overview-fig} shows a hybrid problem of this kind.
We apply a similar pipeline as in program filtering to ensure the blending procedure preserves all the original information.

To further test LLMs on solving hybrid problems of different levels of difficulties.
We create hybrid problems with the following configurations:
(1) \textbf{different numbers and types of options}: let \ttt{G}, \ttt{F}, \ttt{R} denote one option from GSM8K, FOLIO, and ReClor, we create problems of 5 levels of difficulties, that is \ttt{GG}, \ttt{GF}, \ttt{GFX}, \ttt{GFR}, and \ttt{GFRX}.
For example, a \ttt{GG} problem is blended from two GSM8K problems with two options;
a \ttt{GFRX} problem is blended from four problems with the first three each from \ttt{G}, \ttt{F} and \ttt{R}, and \ttt{X} means an option randomly sampled from any of the preceding datasets (Note that \ttt{GFX} only samples from \ttt{G} and \ttt{F}).
We exclude proScript as its answer is a graph and is thus difficult to fit as an option;
and (2) \textbf{different blending granularities}: for each level of problems, we have half of them blended by only adding limited transition words and relevant context is generally put together, while the other half is blended with their sentences shuffled and interleaved together to make it difficult to extract the relevant context (see an example in Table~\ref{tab:hyb-problems} of \S\ref{app:data-gen-examples}).

\textbf{Routing Trajectories}.
Creating routing trajectories is straightforward with the hybrid problems. Since we know the ground-truth original problem associated with each option in the problem, we create corresponding trajectories with each step ``reversing'' the blending process by pasting the original problem as the action and the ground-truth answer as the observation.
Example routing trajectory and hybrid problems shown in \S\ref{app:data-gen-examples}.

The final \dataset contains a total of 6.7K trajectories and 21.7M tokens. Statistics and details are available at \S\ref{app:data-gen-examples} and Table~\ref{tab:data-stats}.







\section{Tactic-Guided Reasoner Fine-Tuning}
\label{sec:tiger-training}
\vspace{-.1cm}

Trajectories provided by \dataset make it possible to fine-tune local LLMs to perform the reasoning in the wild task. However, training on such trajectories is nontrivial and we discuss two different ways to train a LLaMA3-8B model. We refer to these fine-tuned models as \textbf{T}act\textbf{I}c-\textbf{G}uided Reason\textbf{ER} (\method).

\textbf{Learning from imperfect trajectories}.
Trajectories collected in \dataset, while solving the problems, do not always have optimal steps.
For example, a trajectory may involve writing a bad program in step 1 and correcting it with \ttt{Revise Code} action in step 2.
In this case, training the model on predicting tokens of step 1 is improper, as it effectively encourages the model to ``imitate'' a non-optimal step.
Theoretically, a principled solution is to bring in the reinforcement learning technique, as it resembles off-policy learning.
Here, we introduce a simple yet effective approach to the issue while still enjoying the low variance and computational cost of a standard auto-regressive objective.

\begin{figure*}[t]
    \begin{minipage}{\linewidth}
    \begin{minipage}[b]{\linewidth}
        \centering
        \includegraphics[trim={5.1cm 1.8cm 4.4cm .1cm},clip,width=\textwidth]{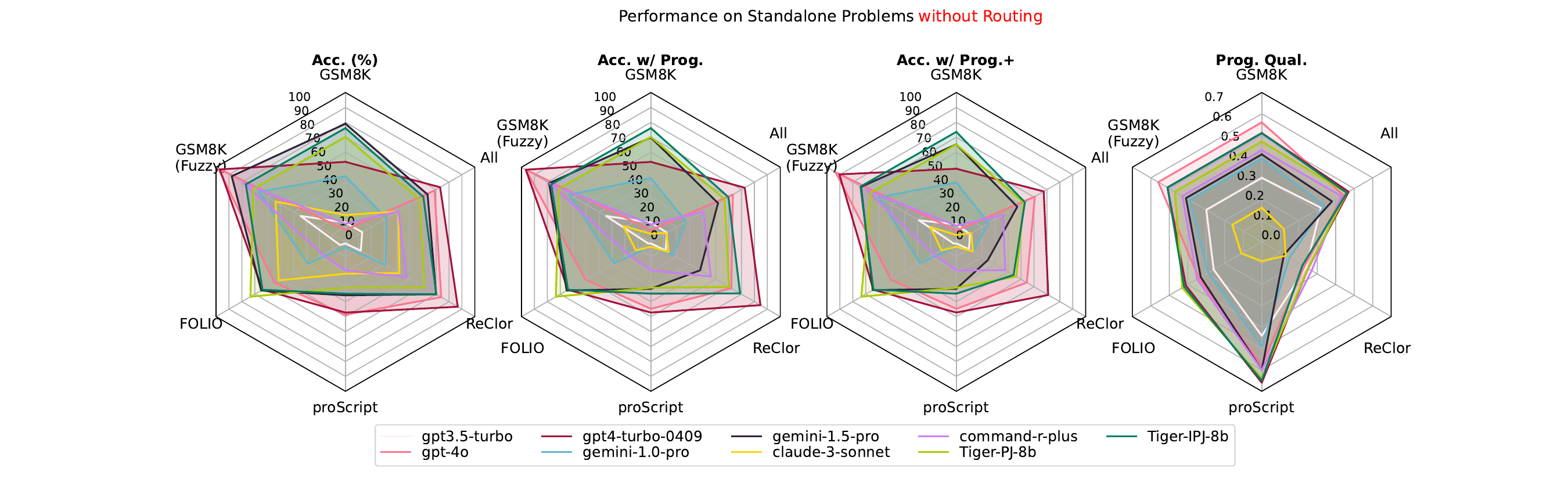}
    \end{minipage}
    \begin{minipage}[b]{\linewidth}
        \centering
        \includegraphics[trim={5.1cm .3cm 4.4cm .1cm},clip,width=\textwidth]{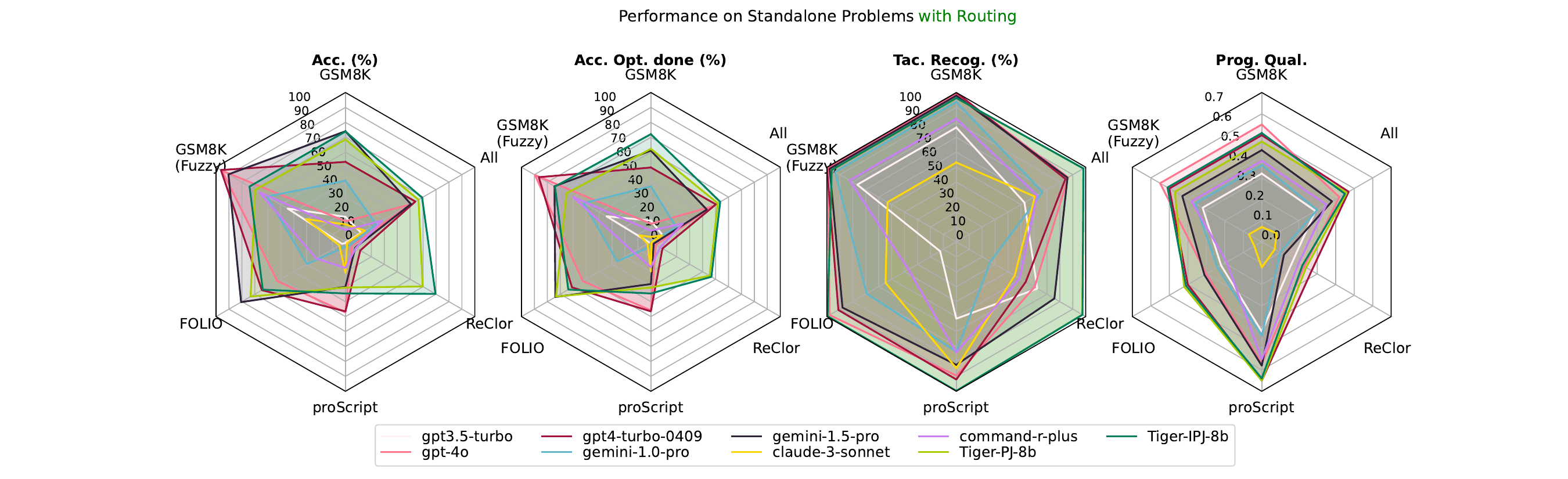}
    \end{minipage}
    \captionof{figure}{Results of the standalone problems. Exact scores provided in \S\ref{app:exp-examples}.}
    \label{fig:standalone-norouting-fig}
\end{minipage}
\end{figure*}

Our approach takes advantage of the fact that ``a bad step is easy to detect in hindsight''.
Consider a trajectory with one bad step
$[\anB{t,a,o}_{c1}, \anB{t,a,o}_{w2}, \anB{t,a,o}_{c3}]$.
Since the LLMs can perform the same action multiple times and some actions are dependent on other actions (e.g., \ttt{Revise Code} can only happen after \ttt{Write Program}), we can easily detect non-optimal steps with rule-based labeling. For example, if \ttt{Revise Code} is performed then any programs produced before it are incorrect; and if \ttt{Answer} is performed then any \ttt{Answer} before this action is incorrect.
Once they are recognized, we experiment with two different approaches:
(1) imperfect trajectory training (\textbf{IPJ}): we mask out the label of $t$ and $a$ of incorrect steps but still train on the original trajectory, in which case the trainable tokens are
$[\anB{t,a,o}_{c1}, o_{w2}, \anB{t,a,o}_{c3}]$;
(2) perfect trajectory training (\textbf{PJ}): we recreate a perfect trajectory by replacing the incorrect ones with subsequent correct ones, in which case a new trajectory
$[\anB{t,a,o}_{c1}, \anB{t_{w2},a_{c3},o_{c3}}]$ is created and trained (assuming $a_{w2}, a_{c3}$ are same type).
Training the model with PJ and IPJ approach is straightforward and we leave details in \S\ref{app:tiger-training-details}.

\vspace{-.1cm}
\section{Experiments}

\begin{figure*}[t]
    \centering
    \includegraphics[trim={4.1cm 0cm 4.5cm .7cm},clip,width=\textwidth]{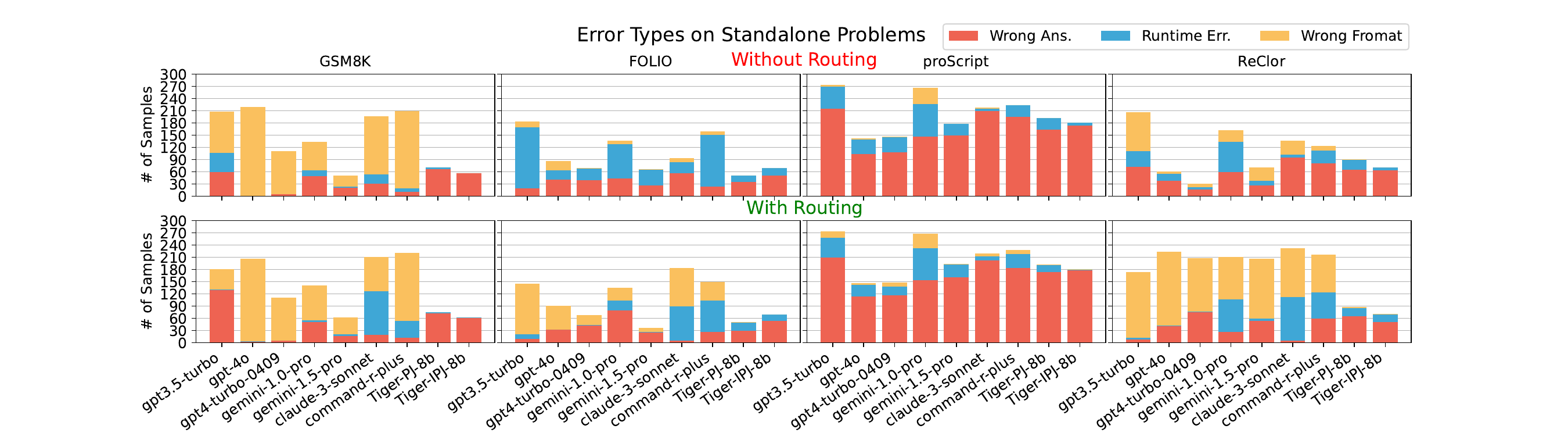}
    \caption{Error types of the standalone problems.}
    \label{fig:error-types-fig}
\end{figure*}
\vspace{-.2cm}



\textbf{Settings}.
We evaluate seven most powerful API LLMs to date listed in Figure~\ref{fig:standalone-norouting-fig} (excluding \ttt{claude-3-opus} due to its high API cost),
and two of our models \methodhead-PJ-8B and \methodhead-IPJ-8B on the standalone and hybrid problems of \dataset.
All API LLMs are given up to five trajectories as ICL prompts (detailed settings in \S\ref{app:exp-examples}).

\subsection{Standalone problems}

We first evaluate models on standalone problems that require only one formalism to solve. Specifically, we evaluate them in two settings:
(1) \textbf{Without routing}: model is given the ground-truth problem-solving tactic, and tasked only to solve the standalone problem, 
and (2) \textbf{with routing}: model is tasked to first find the right tactic and then solve the problem. This is effectively a special case of hybrid problems with only one option.
This can be seen as an easy hybrid benchmark that sits between the two types of problems.

\vspace{-.03cm}
We evaluate the model with the following metrics:
(1) accuracy (\textbf{Acc}): the standard accuracy of models solving the problems;
(2) Acc with program (\textbf{Acc w/ Prog}): a more strict notion of the \textbf{Acc} where trajectories that do not produce a program are considered as failed. This score reduces if the model fails to follow instructions to write the program before answering;
(3) program quality (\textbf{Prog Qual}): recall the ``trivial program'' issue in \S\ref{sec:data-gen}, LLMs can generate a trivial program to shortcut the process. To quantitatively measure this behavior, we measure the difference between the generated program and the ground-truth one using CodeBleu~\citep{ren2020codebleu}, a BLEU score designed for code;
(4) Acc w/ non-trivial program (\textbf{Acc w/ Prog+}): is a more strict version of \textbf{Acc w/ Prog} where the trajectories with programs CodeBleu scores lower than a threshold ($=0.15$) are considered as failed. This metric decreases if a model shortcuts the process and generates trivial programs; 
(5) tactic recognition (\textbf{Tac Recog}): the accuracy of the model correctly recognizes the right tactic for the problem (with routing only);
(6) Acc w/ options done (\textbf{Acc Opt done}): the most strict accuracy where only trajectories that (a) correctly recognized the tactics for all options and (b) generated a program with CodeBleu higher than the threshold for all options, are considered correct. The model having the highest \textbf{Acc Opt done} is the most desired one of this task (with routing only).

\begin{figure*}[t]
    \centering
    \includegraphics[trim={5.35cm 2.7cm 4cm .1cm},clip,width=\textwidth]{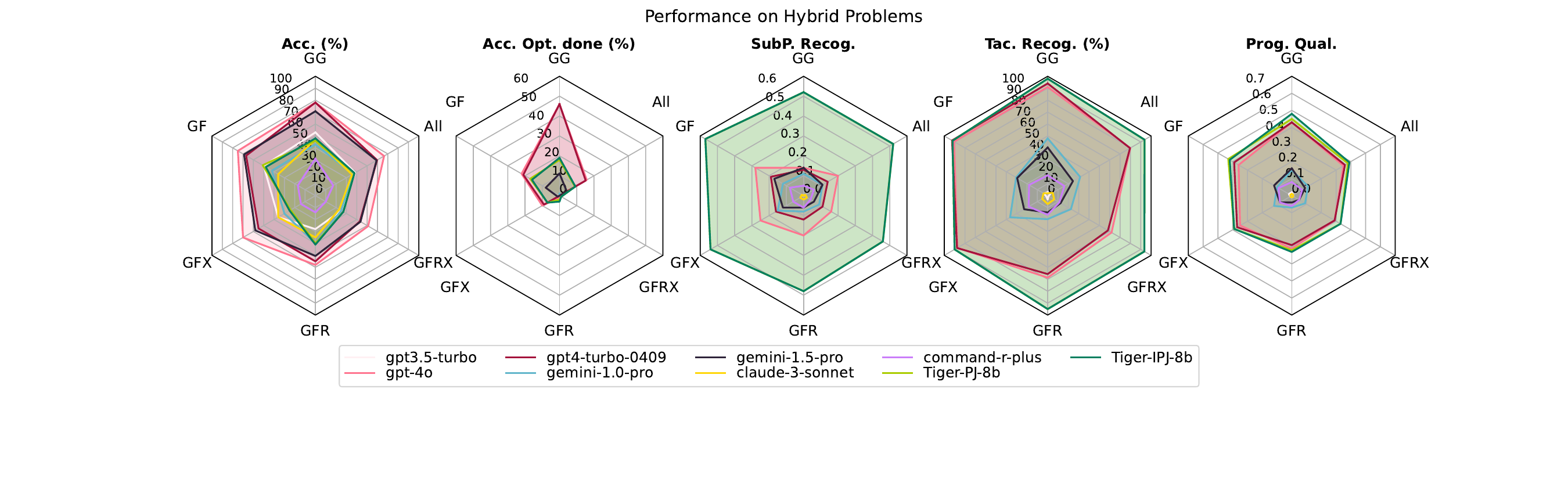}
    \caption{Results of the hybrid problems. Exact scores provided in \S\ref{app:exp-examples}.}
    \label{fig:hyb-fig}    
    \vspace{-.2cm}
\end{figure*}

\textbf{LLMs overfit on popular problems}.
Figure~\ref{fig:standalone-norouting-fig} shows the results with and without the routing.
One major issue we find is that many LLMs while performing well on GSM8K in ICL-CoT, fail significantly in our benchmark, where \ttt{gpt-4o} achieves only 7.95\% on Acc and \ttt{gpt4-turbo-0409} achieves only 53.56\%.
Upon further investigation, we find LLMs tend to ignore our instructions and ICL examples on only returning the numerical answer and proceed to return answers with explanations.
To validate our observation, we categorize the errors into three types and summarize the per-dataset counts in Figure~\ref{fig:error-types-fig}.
Here, \ttt{Wrong Ans} means a valid answer is returned but is incorrect;
\ttt{Runtime Err} means \ttt{Answer} action is never called;
and \ttt{Wrong Format} means \ttt{Answer} is called but the format is invalid, e.g., included irrelevant context and explanations.
Figure~\ref{fig:error-types-fig} shows that all LLMs have exceptionally high \ttt{Wrong Format} rates on GSM8K in both benchmarks; such a phenomenon does not exist for other datasets, precluding the possibility of bad prompting.
Given that the GSM8K dataset comes with CoT ground-truth answers, we conjecture that \textbf{most of the LLMs are trained on these CoT answers which leads to an overfitted behavior in answering}.
Examples shown in \S\ref{app:exp-examples}.

Nevertheless, we implement a fuzzy match pipeline that considers the answer correct as long as the ground-truth numerical answer exists in the answer output. The fuzzy match results are shown under \textbf{GSM8K (fuzzy)}, which aligns much better with that of ICL-CoT, further confirming our observation.
For the hybrid evaluation in \S\ref{sec:hybrid-exp}, we use only the fuzzy match on GSM8K options.

\textbf{Many LLMs lack instruction-following capability in long context}.
We find many LLMs struggle to follow the instructions in a long context. The most prominent case is with \ttt{claude-3-sonnet}, whose overall Acc decreases from 40.83\% to 12.09\% on Acc w/ Prog, indicating the model very frequently skips writing programs before answering.
Other models generally see a decrease of 5\%-10\%, except for \ttt{gpt4-turbo-0409}, which is less than 1\%.

\textbf{Trivial programs and hallucination}.
Figure~\ref{fig:standalone-norouting-fig} sees a general decrease of 10\%-45\% on ReClor problems with Acc w/ Prog+, with the largest drop being 45.73\% from \ttt{gemini-1.5-pro}.
On top of this, the Acc Opt done scores drop to near zero for with-routing benchmarks, indicating a high error rate in recognizing the tactic for ReClor problems. 
This confirms our observation in \S\ref{sec:data-gen} that \textbf{existing LLMs generalize poorly to rarely trained coding problems}.
This observation is further validated in Prog Qual chart, where the CodeBleu scores of ReClor are the lowest.

\textbf{Potential training data imbalance}.
Looking at Figure~\ref{fig:error-types-fig}, we find a high \ttt{Runtime Err} rate with FOLIO, which primarily relies on the z3 library for building the logic model.
We find this is because models frequently write syntactically incorrect code leading to runtime errors and cannot correct it despite multiple \ttt{Revise Code} attempts.
This suggests that the training data likely leans toward popular libraries and the models struggle to generalize to rare libraries.

\textbf{Fine-tuning alleviates above issues}.
\method generally performs better for these issues, with the IPJ version performing slightly better than the PJ one.
\method has more consistent Acc scores for all metrics and maintains the highest score on ReClor in the with-routing setting.
\method's overall performance is similar to GPT4 models, and we believe it would further improve if given more data and trained with larger models.

\subsection{Hybrid problems}
\label{sec:hybrid-exp}

We evaluate LLMs on hybrid problems using a similar set of metrics: note that \textbf{Acc Opt done} here counts the trajectories that answer correctly the hybrid problem with all its options correctly recognized and solved with high-quality programs.
We also include the subproblem recognition metric (\textbf{SubP Recog}) measuring the BLEU score between the extracted subproblem text and the original problem text.

\begin{table}[t]
\centering
\resizebox{0.9\columnwidth}{!}{%
\begin{tabular}{@{}lccccccc@{}}
\toprule
\multicolumn{1}{c}{\multirow{2}{*}{Model}} & \multicolumn{3}{c}{Blend easy}                                                                                                                                                 & \multicolumn{1}{l}{} & \multicolumn{3}{c}{Blend hard}                                                                                                                                                 \\ \cmidrule(lr){2-4} \cmidrule(l){6-8} 
\multicolumn{1}{c}{}                       & \begin{tabular}[c]{@{}c@{}}Acc. Opt.\\ done (\%)\end{tabular} & \begin{tabular}[c]{@{}c@{}}SubP.\\ Recog.\end{tabular} & \begin{tabular}[c]{@{}c@{}}Tac.\\ Recog.\end{tabular} & \multicolumn{1}{l}{} & \begin{tabular}[c]{@{}c@{}}Acc. Opt.\\ done (\%)\end{tabular} & \begin{tabular}[c]{@{}c@{}}SubP.\\ Recog.\end{tabular} & \begin{tabular}[c]{@{}c@{}}Tac.\\ Recog.\end{tabular} \\ \cmidrule(r){1-4} \cmidrule(l){6-8} 
gpt-4o                                     & 16.60                                                         & 0.22                                                   & 81.40                                                 & \multicolumn{1}{l}{} & 14.94                                                         & 0.19                                                   & 78.18                                                 \\
gpt4-turbo-0409                            & 15.83                                                         & 0.15                                                   & 80.69                                                 & \multicolumn{1}{l}{} & 14.52                                                         & 0.14                                                   & 78.46                                                 \\
gemini-1.5-pro                             & 4.53                                                          & 0.12                                                   & 26.1                                                  &                      & 3.93                                                          & 0.1                                                    & 22.63                                                 
\\ \cmidrule(r){1-4} \cmidrule(l){6-8} 
Tiger-PJ-8b                                & 8.94                                                          & 0.52                                                   & 93.02                                                 &                      & 7.76                                                          & 0.5                                                    & 94.43                                                 \\
Tiger-IPJ-8b                               & 8.88                                                          & 0.52                                                   & 93.02                                                 &                      & 9.54                                                          & 0.5                                                    & 4.43                                                  \\ \bottomrule
\end{tabular}%
}
\caption{Results of hybrid problems with different blending strategies (zero-scored model removed).}
\label{tab:blend-diff-table}
\vspace{-.05cm}
\end{table}

\textbf{Hybrid problems are highly challenging}.
We summarize the results in Figure~\ref{fig:hyb-fig}.
All models maintain an Acc score above 35\% on hybrid problems except for \ttt{command-r-plus}.
We observe a consistent drop in scores of all models when moving to problems of higher difficulties, indicating that difficulty levels are effective.
Remarkably, the Acc Opt done score drops significantly, where most LLMs drop to near zero, with only GPT series and \method maintaining an overall score of around 10\%.
Furthermore, most of the scores are earned on easy ones such as \ttt{GG}, and all API LLMs could not even successfully answer a single problem from \ttt{GFR} and \ttt{GFRX}.
Thanks to fine-tuning, \method maintained the highest SubP and Tac Recog scores and the best Prog Qual, and managed to solve a few more problems in hard difficulties.
We conjecture the task has gone beyond the capacity of an 8B model and we might see a bigger gap by fine-tuning larger models.
Comparing all LLMs, GPT4 series has a clear advantage, still, their capabilities are far from obtaining a reasonable score in this benchmark.
In summary, \textbf{we believe the hybrid problems remain a highly challenging and valuable benchmark that provides deep insights into LLMs' reasoning capability}.

\textbf{Blending strategies}.
We summarize the results on different blending strategies in 
Table~\ref{tab:blend-diff-table}.
We observe a drop in the Acc and SubP Recog moving from easy to hard for many LLMs, but they are not overall very sensitive to different blending strategies.
This suggests that LLMs are generally strong at retrieval in short context.
We leave the investigation in long context in future work.





\vspace{-.1cm}
\section{Conclusion}
\vspace{-.1cm}
We present \textit{reasoning in the wild}, a task that challenges LLMs' reasoning capabilities in solving ambiguous and mixed-in-scope problems, where we create dataset \dataset.
Our experiments reveal critical issues of the existing LLMs and show that they can be alleviated with fine-tuning, that is \method.

\clearpage
\newpage

\section{Limitations}

During the creation of \dataset, we used a combined approach of manual labeling and auto filtering.
It is likely the trajectories, especially for ReClor data, still contain ``trivial programs''.

During evaluation, we use two different ICL prompts for problem-solving trajectories to balance between cost and performance.
While the total length of the message can go to 10K long, it does not exhaust all the context window length.
That said, it is likely by providing more fully detailed trajectories, the API LLMs can achieve better performance than those reported in the paper.

\section{Ethics Statement}

\textbf{Potential negative impact}.
The trajectory data contain fully the LLM-generated contents, and the code and text could potentially contain misinformation.
It is also possible, while unlikely that running the programs in the trajectory could cause unexpected behaviors in the host machine due to different environment configurations.

\textbf{Artifact statements}.
We release data, code, and weights under Apache 2.0 license and they are intended for research use only. Additionally, the usage of the \method model should follow the license agreement of LLaMA and Alpaca.
\dataset is released under CC BY NC 4.0 and the use of such dataset should follow the \href{https://openai.com/policies/terms-of-use}{policy of OpenAI}.


\bibliography{custom}


\appendix

\section{Reasoning in the Wild examples}
\label{app:method-examples}

\textbf{Tactic description examples}.
An example of a complete tactic description is shown in Table~\ref{tab:tactic-doc}.

\section{Data generation}
\label{app:data-gen-examples}

\textbf{Tactic Pool}.
We manually create a pool of diverse tactics for different types of reasoning including:
(1) \textbf{Math Tactic}, where the agent builds a math model to solve problems involving math;
(2) \textbf{Logic Tactic}, where the agent builds a logic model using the Python Z3 library to model and verify logical statements, which can be used for logical deduction, induction, and abduction problems;
(3) \textbf{Graph Tactic}, where the agent builds a graph model using the Python NetworkX library to model a graph structure, which can be used for problems involving sorting items as a graph;
(4) \textbf{General Program Tactic}, where the agent writes a general program using all libraries above to represent problems that do not fit into any of the formalisms above, such as commonsense reasoning;
and (5) \textbf{Routing Tactic}, which is the main tactic, which the agent follows to decompose the problem into subproblems and identify the best tactic for solving it.

\textbf{Datasets}.
We select the following datasets each corresponding to our designed tactics above for trajectory generation: 
(1) GSM8K~\citep{cobbe2021training}: a popular arithmetic 
math dataset;
(2) FOLIO~\citep{han2022folio}: a logic-grounded natural language inference (NLI) dataset;
(3) ProScript~\citep{sakaguchi2021proscript}: a dataset containing graph-structured natural language steps for achieving certain goals such as ``opening a bank account'';
and (4) ReClor~\citep{yu2020reclor}: a commonsense reasoning dataset curated from LSAT and GMAT tests in the form of multichoice problems.

\textbf{Trajectory collection process}.
To generate trajectories that adhere to our setup, we first manually created trajectories for each dataset, then include them as ICL prompts to guide the generation.
We use a mixture of three models in the process:
\ttt{gpt4-turbo-0409}, \ttt{gpt-4o}, and \ttt{claude-3-opus}.
In particular, we find \ttt{claude-3-opus} performs the best in zero-shot and out-of-distribution problems such as ReClor, and GPT4 series have the best instruction-following capability during generation.
Based on our initial experiments, we set the maximum number of steps of each trajectory to 7 to balance between the final success rate and the cost.
A trajectory is terminated if (1) the \ttt{Answer} action is called and the result is correct; (2) the max steps are reached; (3) the model makes 3 consecutive errors, which is typically a bad sign of a failed trajectory.

\textbf{Post-processing}.
After generation, we filter the trajectories with the following steps:
(1) We filter those that do not write any programs or those with programs that failed to run.
This indicates the model ``shortcuts'' the process by directly outputting the answer.
(2) A more severe shortcutting happens with ReClor dataset for virtually all the LLMs we tested.
We refer to this as ``\textbf{trivial programs}'':
ReClor problems are typically ambiguous in scope and do not fit into any existing formalisms are known to be trained by these LLMs, such as math and formal logic.
When LLMs are ``forced'' to generate programs for them, even with multiple ICL prompts given, they very often generate programs that either
(a) are a fancy CoT, where they put CoT free-form reasoning in the comments and have programs directly return the answer;
(b) or have the answer ``hardcoded'' as constants.
Examples are shown in Table~\ref{tab:trivial-prog}.
We filter these trajectories by first manually labeling a subset of good and bad programs, and use LLMs to filter the rest using them as ICL prompts.
To further boost the filtering accuracy, we use a majority vote of three LLMs: \ttt{gpt4-turbo-0409}, \ttt{claude-3-opus}, and \ttt{gemini-1.5-pro}.
On our manually labeled test set, this framework achieves an accuracy of 72\% with 16\% being false positives (i.e., the bad programs inferred as good ones).
We measure Krippendorff's alpha, that is the inter-model agreement rate, which is 0.69, indicating the annotations are generally aligned.
In experiments, we show that this phenomenon is widely present in all LLMs and causes a significant performance drop in all ReClor-related problems.

\textbf{Trivial program examples}.
Trivial program examples shown in Table~\ref{tab:trivial-prog}.

\textbf{Hybrid problem examples}.
Hybrid problem examples shown in Table~\ref{tab:hyb-problems}.

\textbf{Routing trajectory examples}.
Routing trajectory examples shown in Table~\ref{tab:rout-traj}.

\begin{table}[t]
\centering
\resizebox{0.9\columnwidth}{!}{%
\begin{tabular}{@{}lcccc@{}}
\toprule
\multicolumn{1}{c}{Subset} & \# Train & \# Test & Avg. steps & Avg. Tokens \\ \midrule
GSM8K                      & 1.2K     & 239     & 4.2        & 2.1K        \\
FOLIO                      & 0.9K     & 192     & 4.3        & 4.6K        \\
proScript                  & 1.1K     & 277     & 5          & 3.4K        \\
ReClor                     & 1.1K     & 235     & 5.2        & 4.3K        \\
Hybrid                     & 1K       & 500     & 4          & 2K          \\ \midrule
All                        & 5.3K     & 1.4K    & 4.5        & 3.2K        \\ \bottomrule
\end{tabular}%
}
\caption{\dataset dataset statistics.}
\label{tab:data-stats}
\end{table}

\textbf{Dataset statistics}.
We show statistics of~\dataset in Table~\ref{tab:data-stats}.
Due to budget limits, we did not generate trajectories for all the problems. We plan to release the generation pipeline for researchers interested in continuing the process.
For each dataset and the hybrid one, we aim for 1K for training and 250 for testing (FOLIO has fewer total testing samples).
\dataset contains a total of 6.7K trajectories and 21.7M tokens.
In experiments, we find it supports a sufficient fine-tuning of an LLaMA3-8B model and evaluation of existing LLMs.

\section{Tactic-Guided Reasoner Fine-Tuning}
\label{app:tiger-training-details}

In \S\ref{sec:tiger-training}, we introduce two ways to prepare the trajectory data: perfect trajectory training (PJ) and imperfect trajectory training (IPJ).
PJ training is straightforward as it resembles standard imitation learning and is applied in prior work such as ReAct~\citep{yao2022react} and ToRA~\citep{gou2023tora}.
We find IPJ training could further improve the model's performance because it trains the model to correct the previous steps, making it more robust during inference time.
In experiments, we fine-tune two versions of LLaMA3-8B:
(1) \methodhead-PJ-8B which trains on PJ data;
(2) \methodhead-IPJ-8B which trains on PJ and IPJ data.
Both models are trained with LoRA $r=64, \alpha=64$.
Additionally, we train the routing model by continuing training on the \methodhead-PJ-8B on routing trajectories with LoRA $r=16, \alpha=16$.
All training is performed for 3 epochs on a single A100 GPU.

\section{Experiments}
\label{app:exp-examples}

\textbf{Hyperparameters}.
For all experiments, We set the max steps allowed in a trajectory to 7, and the max consecutive errors to 3.
For routing trajectory, all API LLMs are given two full routing trajectories as ICL prompts.
For problem-solving trajectory, all API LLMs are provided with 5-shot ICL prompts consisting of the first two steps (typically, \ttt{Plan} and \ttt{Write Program}) to a problem.
This prompt is given for the first two steps of the in-progress trajectory, then we swap it into ICL prompts of two problem-solving trajectories of the same problem type with their main output contents omitted.
We do this because we find the most difficult part of a trajectory is the first few steps, where the model figures out the right overall approach, so we insert more detailed solutions. And for the rest, we insert a complete trajectory to guide the model with formatting.
This helps to balance between the performance and cost as one trajectory ICL example is 3K long.

For CodeBleu metric~\citep{ren2020codebleu}, we set the weights of \ttt{ngram match}, \ttt{weighted ngram match}, \ttt{syntax match}, \ttt{dataflow match} to $0.15, 0.15,0.35,0.35$ to favor programs with similar functionality rather than textual appearance;

\textbf{GSM8K overfitted answers}.
Example overfitted outputs of GSM8K shown in Table~\ref{tab:wrong-format}.

\begin{table*}[t]
\centering
\resizebox{\textwidth}{!}{%
\begin{tabular}{@{}ll@{}}
\toprule
\multicolumn{1}{c}{gpt4-turbo-0409 Output}                                                                                                                                                                                                                                                                                                                                      & \multicolumn{1}{c}{gemini-1.5-pro Output}                                                                                                                                                                                                                                        \\ \midrule
\begin{tabular}[c]{@{}l@{}}out\_content:\\ The rainfall on Tuesday is 5 inches.\\ gt answer:\\ 5\\ \\ out\_content:\\ The monthly earnings of the dance studio are a net loss of \$120.00.\\ gt answer:\\ 480\\ \\ out\_content:\\ James made a total commission of \$17,500 from selling 10 cars.\\ gt answer:\\ 17500\end{tabular}                                            & \begin{tabular}[c]{@{}l@{}}out\_content:\\ Dorothy has 4 Facebook friends. \\ James has 16 Facebook friends.\\ gt answer:\\ 16\\ \\ \\ out\_content:\\ 5\\ ===\\ gt answer:\\ 12\\ \\ out\_content:\\ 42.0\\ ===\\ gt answer:\\ 1248\end{tabular}                                \\ \midrule
\multicolumn{1}{c}{claude-3-sonnet Output}                                                                                                                                                                                                                                                                                                                                      & \multicolumn{1}{c}{command-r-plus Output}                                                                                                                                                                                                                                        \\ \midrule
\begin{tabular}[c]{@{}l@{}}out\_content:\\ The number of pieces of candy that are not chocolate is 49.\\ gt answer:\\ 20\\ \\ out\_content:\\ The total dollar amount in a stack containing two thirds of the 9,300 pennies is \$62.00.\\ gt answer:\\ 62\\ \\ out\_content:\\ The total amount Molly will pay for catering the party is \$131.\\ gt answer:\\ 101\end{tabular} & \begin{tabular}[c]{@{}l@{}}out\_content:\\ The answer is **40**.\\ gt answer:\\ 140\\ \\ out\_content:\\ The answer to the question is **180000 meters**.\\ gt answer:\\ 180000\\ \\ out\_content:\\ There are 12 teachers at Dr. Wertz's school.\\ gt answer:\\ 36\end{tabular} \\ \bottomrule
\end{tabular}%
}
\caption{Example of overfitted outputs on GSM8K problems.}
\label{tab:wrong-format}
\end{table*}

\textbf{Full results}.
We show the numerical results in Table~\ref{tab:res-table-norout}, Table~\ref{tab:res-table-rout}, and Table~\ref{tab:res-table-hyb}.

\begin{table*}[t]
\centering
\resizebox{\textwidth}{!}{%
\begin{tabular}{@{}lcccccccccccc@{}}
\toprule
\multicolumn{1}{c}{\multirow{2}{*}{Model}} & \multicolumn{4}{c}{GSM8K}                            & \multicolumn{4}{c}{GSM8K Soft}                       & \multicolumn{4}{c}{FOLIO}                            \\
\multicolumn{1}{c}{}                       & Acc (\%) & Acc w/ Prog. & Acc w/ Prog+ & Prog. Qual. & Acc (\%) & Acc w/ Prog. & Acc w/ Prog+ & Prog. Qual. & Acc (\%) & Acc w/ Prog. & Acc w/ Prog+ & Prog. Qual. \\ \midrule
gpt3.5-turbo                               & 12.97    & 12.97        & 11.72        & 0.3         & 34.73    & 34.73        & 28.87        & 0.3         & 4.17     & 2.08         & 2.08         & 0.26        \\
gpt-4o                                     & 7.95     & 7.95         & 7.53         & 0.56        & 98.33    & 96.65        & 93.31        & 0.56        & 54.69    & 50.52        & 50.52        & 0.34        \\
gpt4-turbo-0409                            & 53.56    & 53.56        & 48.95        & 0.51        & 97.07    & 96.65        & 90.38        & 0.51        & 64.06    & 64.06        & 64.06        & 0.41        \\
gemini-1.0-pro                             & 43.93    & 42.68        & 39.75        & 0.39        & 66.95    & 63.18        & 59.83        & 0.39        & 29.17    & 28.65        & 28.12        & 0.29        \\
gemini-1.5-pro                             & 79.08    & 69.87        & 65.27        & 0.41        & 87.87    & 78.66        & 73.64        & 0.41        & 65.1     & 65.1         & 64.58        & 0.33        \\
claude-3-sonnet                            & 17.99    & 5.44         & 4.6          & 0.16        & 53.97    & 21.34        & 20.08        & 0.16        & 51.56    & 11.46        & 11.46        & 0.11        \\
command-r-plus                             & 12.13    & 11.72        & 10.88        & 0.43        & 79.08    & 78.24        & 69.46        & 0.43        & 17.19    & 17.19        & 17.19        & 0.35        \\
Tiger-PJ-8b                                & 70.29    & 70.29        & 65.27        & 0.47        & 71.13    & 71.13        & 66.11        & 0.47        & 73.44    & 73.44        & 73.44        & 0.43        \\
Tiger-IPJ-8b                               & 76.15    & 76.15        & 73.64        & 0.51        & 76.99    & 76.99        & 74.06        & 0.51        & 64.06    & 64.06        & 64.06        & 0.42        \\ \midrule
\multicolumn{1}{c}{\multirow{2}{*}{Model}} & \multicolumn{4}{c}{proScript}                        & \multicolumn{4}{c}{ReClor}                           & \multicolumn{4}{c}{All}                              \\
\multicolumn{1}{c}{}                       & Acc (\%) & Acc w/ Prog. & Acc w/ Prog+ & Prog. Qual. & Acc (\%) & Acc w/ Prog. & Acc w/ Prog+ & Prog. Qual. & Acc (\%) & Acc w/ Prog. & Acc w/ Prog+ & Prog. Qual. \\ \midrule
gpt3.5-turbo                               & 1.08     & 1.08         & 1.08         & 0.44        & 11.91    & 11.49        & 9.79         & 0.24        & 12.94    & 12.41        & 10.5         & 0.32        \\
gpt-4o                                     & 48.74    & 44.77        & 44.77        & 0.58        & 74.04    & 62.13        & 54.47        & 0.23        & 68.82    & 63.41        & 60.66        & 0.44        \\
gpt4-turbo-0409                            & 47.29    & 47.29        & 47.29        & 0.66        & 86.81    & 84.68        & 71.06        & 0.25        & 73.17    & 72.53        & 67.55        & 0.47        \\
gemini-1.0-pro                             & 3.61     & 3.25         & 3.25         & 0.49        & 31.06    & 17.45        & 12.77        & 0.15        & 31.71    & 27.15        & 25.03        & 0.34        \\
gemini-1.5-pro                             & 35.74    & 31.41        & 31.41        & 0.6         & 70.09    & 38.03        & 24.36        & 0.12        & 63.48    & 51.91        & 47.13        & 0.38        \\
claude-3-sonnet                            & 21.3     & 3.25         & 3.25         & 0.09        & 41.7     & 13.62        & 12.77        & 0.13        & 40.83    & 12.09        & 11.56        & 0.12        \\
command-r-plus                             & 19.13    & 19.13        & 19.13        & 0.6         & 47.23    & 46.38        & 37.87        & 0.27        & 40.93    & 40.51        & 36.16        & 0.43        \\
Tiger-PJ-8b                                & 30.69    & 30.69        & 30.69        & 0.64        & 61.28    & 60.43        & 46.38        & 0.25        & 57.26    & 57.05        & 52.28        & 0.46        \\
Tiger-IPJ-8b                               & 34.66    & 34.66        & 34.66        & 0.65        & 70.21    & 68.94        & 44.26        & 0.22        & 60.23    & 59.92        & 53.02        & 0.46        \\ \bottomrule
\end{tabular}%
}
\caption{Results of standalone problems without routing.}
\label{tab:res-table-norout}
\end{table*}

\begin{table*}[t]
\centering
\resizebox{\textwidth}{!}{%
\begin{tabular}{@{}lcccccccccccc@{}}
\toprule
\multicolumn{1}{c}{\multirow{2}{*}{Model}} & \multicolumn{4}{c}{GSM8K}                            & \multicolumn{4}{c}{GSM8K Soft}                       & \multicolumn{4}{c}{FOLIO}                            \\
\multicolumn{1}{c}{}                       & Acc (\%) & Acc w/ Prog. & Acc w/ Prog+ & Prog. Qual. & Acc (\%) & Acc w/ Prog. & Acc w/ Prog+ & Prog. Qual. & Acc (\%) & Acc w/ Prog. & Acc w/ Prog+ & Prog. Qual. \\ \midrule
gpt3.5-turbo                               & 16.74    & 14.23        & 76.57        & 0.32        & 44.77    & 34.31        & 76.57        & 0.32        & 2.6      & 1.04         & 12.5         & 0.22        \\
gpt-4o                                     & 13.81    & 12.13        & 99.16        & 0.55        & 95.4     & 89.54        & 99.16        & 0.55        & 52.6     & 52.6         & 97.92        & 0.3         \\
gpt4-turbo-0409                            & 53.56    & 49.79        & 100          & 0.5         & 96.23    & 86.61        & 100          & 0.5         & 64.58    & 60.94        & 91.15        & 0.4         \\
gemini-1.0-pro                             & 41       & 37.24        & 93.72        & 0.36        & 60.67    & 51.88        & 93.72        & 0.36        & 29.69    & 26.04        & 69.27        & 0.24        \\
gemini-1.5-pro                             & 74.06    & 61.09        & 97.91        & 0.43        & 90.38    & 74.48        & 97.91        & 0.43        & 80.63    & 73.82        & 87.96        & 0.31        \\
claude-3-sonnet                            & 11.72    & 3.35         & 53.14        & 0.07        & 30.54    & 8.79         & 53.14        & 0.07        & 4.69     & 2.6          & 54.69        & 0.04        \\
command-r-plus                             & 7.95     & 7.53         & 82.43        & 0.38        & 66.53    & 60.25        & 82.43        & 0.38        & 22.4     & 13.54        & 36.98        & 0.19        \\
Tiger-PJ-8b                                & 68.55    & 62.28        & 96.2         & 0.47        & 69.76    & 65.19        & 96.2         & 0.47        & 73.29    & 73.29        & 99.5         & 0.42        \\
Tiger-IPJ-8b                               & 73.9     & 72.11        & 96.3         & 0.51        & 74.13    & 74.12        & 96.3         & 0.51        & 63.89    & 63.84        & 99.5         & 0.41        \\ \midrule
\multicolumn{1}{c}{\multirow{2}{*}{Model}} & \multicolumn{4}{c}{proScript}                        & \multicolumn{4}{c}{ReClor}                           & \multicolumn{4}{c}{All}                              \\
\multicolumn{1}{c}{}                       & Acc (\%) & Acc w/ Prog. & Acc w/ Prog+ & Prog. Qual. & Acc (\%) & Acc w/ Prog. & Acc w/ Prog+ & Prog. Qual. & Acc (\%) & Acc w/ Prog. & Acc w/ Prog+ & Prog. Qual. \\ \midrule
gpt3.5-turbo                               & 0.8      & 0.5          & 51.4         & 0.42        & 0.43     & 0            & 62.55        & 0.17        & 12.22    & 9.05         & 52.64        & 0.29        \\
gpt-4o                                     & 47.2     & 45.8         & 89.52        & 0.57        & 5.11     & 3.83         & 60.43        & 0.23        & 50.03    & 47.81        & 86.42        & 0.42        \\
gpt4-turbo-0409                            & 46.5     & 46.5         & 92.12        & 0.65        & 11.49    & 8.94         & 53.62        & 0.27        & 54.06    & 50.24        & 84.32        & 0.47        \\
gemini-1.0-pro                             & 3.21     & 3.04         & 73.49        & 0.44        & 4.09     & 0            & 26.36        & 0.1         & 23.7     & 19.66        & 66.66        & 0.3         \\
gemini-1.5-pro                             & 30.1     & 28.21        & 82.31        & 0.58        & 7.62     & 2.24         & 75.78        & 0.12        & 50.58    & 43.24        & 85.91        & 0.38        \\
claude-3-sonnet                            & 20.8     & 19.9         & 84.56        & 0.12        & 1.28     & 0.43         & 45.11        & 0.07        & 15.12    & 8.71         & 60.68        & 0.08        \\
command-r-plus                             & 17.4     & 16.9         & 74.12        & 0.55        & 8.09     & 7.23         & 48.94        & 0.2         & 28.55    & 24.79        & 62.39        & 0.35        \\
Tiger-PJ-8b                                & 30.54    & 30.54        & 99.8         & 0.65        & 59.71    & 45.37        & 97.4         & 0.24        & 56.45    & 51.72        & 98.23        & 0.46        \\
Tiger-IPJ-8b                               & 34.53    & 34.53        & 99.7         & 0.64        & 69.57    & 46.77        & 97.5         & 0.22        & 59.24    & 53.58        & 98.25        & 0.45        \\ \bottomrule
\end{tabular}%
}
\caption{Results of standalone problems with routing.}
\label{tab:res-table-rout}
\end{table*}

\begin{table*}[t]
\centering
\resizebox{\textwidth}{!}{%
\begin{tabular}{@{}lccccccccccccccc@{}}
\toprule
\multicolumn{1}{c}{\multirow{2}{*}{Model}} & \multicolumn{5}{c}{gg}                                                   & \multicolumn{5}{c}{gf}                                                   & \multicolumn{5}{c}{gfx}                                                  \\
\multicolumn{1}{c}{}                       & Acc (\%) & Acc. Opt done (\%) & SubP. Recog. & Tac. Recog. & Prog. Qual. & Acc (\%) & Acc. Opt done (\%) & SubP. Recog. & Tac. Recog. & Prog. Qual. & Acc (\%) & Acc. Opt done (\%) & SubP. Recog. & Tac. Recog. & Prog. Qual. \\ \midrule
gpt3.5-turbo                               & 53       & 0                  & 0            & 3           & 0.02        & 50       & 0                  & 0.01         & 3.5         & 0.02        & 36       & 0                  & 0            & 1.33        & 0.01        \\
gpt-4o                                     & 78       & 46                 & 0.14         & 91          & 0.43        & 75       & 22                 & 0.28         & 90.5        & 0.36        & 70       & 10                 & 0.25         & 87.33       & 0.35        \\
gpt4-turbo-0409                            & 78       & 46                 & 0.13         & 94          & 0.43        & 67       & 21                 & 0.19         & 92.5        & 0.39        & 55       & 9                  & 0.16         & 87.67       & 0.37        \\
gemini-1.0-pro                             & 43       & 1                  & 0.11         & 48          & 0.15        & 42       & 0                  & 0.12         & 30.5        & 0.1         & 30       & 0                  & 0.15         & 36.33       & 0.12        \\
gemini-1.5-pro                             & 71       & 11                 & 0.14         & 40.4        & 0.16        & 69       & 8                  & 0.17         & 29.5        & 0.12        & 58       & 1                  & 0.12         & 22.79       & 0.08        \\
claude-3-sonnet                            & 47       & 0                  & 0            & 2           & 0.01        & 36       & 0                  & 0.01         & 4           & 0.01        & 36       & 0                  & 0.02         & 5.33        & 0.01        \\
command-r-plus                             & 31       & 0                  & 0.05         & 17          & 0.08        & 17       & 0                  & 0.08         & 18          & 0.09        & 14       & 0                  & 0.06         & 18.33       & 0.08        \\
Tiger-PJ-8b                                & 46       & 18                 & 0.52         & 98          & 0.45        & 51       & 17                 & 0.57         & 92          & 0.43        & 24       & 7                  & 0.54         & 90          & 0.39        \\
Tiger-IPJ-8b                               & 48       & 19                 & 0.52         & 98          & 0.48        & 48       & 16                 & 0.57         & 92          & 0.42        & 25       & 7                  & 0.54         & 90          & 0.39        \\ \midrule
\multicolumn{1}{c}{\multirow{2}{*}{Model}} & \multicolumn{5}{c}{gfr}                                                  & \multicolumn{5}{c}{gfrx}                                                 & \multicolumn{5}{c}{all}                                                  \\
\multicolumn{1}{c}{}                       & Acc (\%) & Acc. Opt done (\%) & SubP. Recog. & Tac. Recog. & Prog. Qual. & Acc (\%) & Acc. Opt done (\%) & SubP. Recog. & Tac. Recog. & Prog. Qual. & Acc (\%) & Acc. Opt done (\%) & SubP. Recog. & Tac. Recog. & Prog. Qual. \\ \midrule
gpt3.5-turbo                               & 28       & 0                  & 0            & 3.33        & 0.01        & 25       & 0                  & 0            & 0.75        & 0.01        & 38.4     & 0                  & 0.01         & 2.38        & 0.01        \\
gpt-4o                                     & 58       & 1                  & 0.2          & 68.67       & 0.31        & 51       & 0                  & 0.16         & 61.75       & 0.29        & 66.4     & 15.8               & 0.2          & 79.85       & 0.35        \\
gpt4-turbo-0409                            & 55       & 0                  & 0.12         & 65.67       & 0.29        & 43       & 0                  & 0.11         & 58.25       & 0.29        & 59.6     & 15.2               & 0.14         & 79.62       & 0.36        \\
gemini-1.0-pro                             & 40       & 0                  & 0.08         & 19.67       & 0.07        & 25       & 0                  & 0.09         & 22.5        & 0.09        & 36       & 0.2                & 0.11         & 31.4        & 0.11        \\
gemini-1.5-pro                             & 51       & 0                  & 0.06         & 14.09       & 0.04        & 44       & 0                  & 0.06         & 12.5        & 0.05        & 59.11    & 4                  & 0            & 24.42       & 0.09        \\
claude-3-sonnet                            & 35       & 0                  & 0.02         & 7           & 0.01        & 23       & 0                  & 0.02         & 6.25        & 0.01        & 35.4     & 0                  & 0.01         & 4.92        & 0.01        \\
command-r-plus                             & 14       & 0                  & 0.06         & 16          & 0.06        & 10       & 0                  & 0.04         & 11.25       & 0.05        & 17.2     & 0                  & 0.06         & 16.12       & 0.07        \\
Tiger-PJ-8b                                & 40       & 2                  & 0.48         & 95          & 0.32        & 27       & 1                  & 0.46         & 93.5        & 0.33        & 37.6     & 9                  & 0.52         & 93.7        & 0.38        \\
Tiger-IPJ-8b                               & 41       & 3                  & 0.48         & 95          & 0.33        & 27       & 1                  & 0.46         & 93.5        & 0.33        & 37.8     & 9.2                & 0.52         & 93.7        & 0.39        \\ \bottomrule
\end{tabular}%
}
\caption{Results of hybrid problems.}
\label{tab:res-table-hyb}
\end{table*}


\clearpage
\onecolumn
{\small\tabcolsep=3pt  
\centering
\begin{xtabular}{@{}l@{}}
\toprule
Predicate Logic Tactic
                                                           \\ \midrule
\#\#\# Tactic name\\ predicate\_logic\_z3\\ \\ \#\#\# Problem type and tactic\\ This tactic builds a formal logical model using predicate logic formalism with the help of python z3 lib.\\ This tactic is suitable for solving reasoning problems that involves deductive, inductive or, abductive reasoning.\\ To do so, the tactic will represent the problem as a self-contained first-order logic (FOL) system that consists\\ of Constants, Predicates, Logic Variables, Quantifiers, Functions, Logic Operators, Grounded Facts, Logic Formulas\\ and so on; then it will seek to perform formal reasoning with the help with z3 lib.\\ \\ **Typical use cases**\\ The tactic is suitable for problems that can be represented by an FOL system and solved by performing the following\\ three types of formal reasoning\\ \\ - Deductive reasoning: Given Facts and Logic Formulas, deduce new Facts from the system by applying the Formulas to the\\   Facts.\\ - Inductive reasoning: Given Facts and Potentially some Formulas, induce new Formulas that entail the given Facts and\\   are consistent with the preexisting Formulas.\\ - Abductive reasoning: Given Facts, Logic Formulas, and a consequence Fact, infer the missing Facts or Formulas, such\\   that the consequence Fact can be entailed by the system.\\ \\ **Model and tactic outputs**\\ - Model: To apply the tactic, one builds a self-contained FOL system that fully represent the problem using z3 lib\\ \\ - Outputs: the z3 code should output either 'Agree', 'Contradict', or 'Uncertain'.\\   'Agree' means the Facts or Formulas agree with the system\\   'Contradict' means the Facts or Formulas contradict with the system\\   'Uncertain' means the Facts or Formulas contradict with the system\\ \\ Note that the type of reasoning and the system built for the problem determine:\\  - How the output is interpreted.\\  - Whether the output serves as the final answer or intermediate checks for the problem-specific answer\\ For example: for a deductive reasoning task with a given hypothesis, one builds the system to determine if the\\ hypothesis Agree/Contradict/Uncertain to the system; for a deductive reasoning task where one wants to deduce all\\ possible Facts, then one should infer all Facts that Agree with the system; for inductive reasoning, one infers the\\ Formulas that Agree with the system; for abductive reasoning, one infers the Facts or Formulas that Agree with the\\ consequence and the system.\\ \\ \\ \#\#\# Tactic details\\ You will use the following python libs to solve the problem:\\ Any builtin Python libs\\ z3\\ \\ **Code template**\\ You will use the following code template to solve the problem.\\ \\ ```python\\ import z3\\ from z3 import *\\ \\ def check\_model(solver):\\     res = solver.check()\\     if res == sat:\\         return 'sat'\\     elif res == unsat:\\         return 'unsat'\\     else:\\         return 'unsolvable'\\ \\ def check\_constraint(solver, c):\\     pos\_res = solver.check(c)\\     neg\_res = solver.check(Not(c))\\ \\     if (pos\_res == sat) and (neg\_res == unsat):\\         return 'Agree'\\     elif (pos\_res == unsat) and (neg\_res == sat):\\         return 'Contradict'\\     elif (pos\_res == unknown) or (neg\_res == unknown):\\         return 'unsolvable'\\     else:\\         return 'Uncertain'\\ \\ def main():\\     s = z3.Solver()\\     \textless{}your code\textgreater\\ ```\\ \\ **Action space**\\ You will use and ONLY use the following actions to solve the problem.\\ You can apply actions in arbitrary order and arbitrary number of times.\\ \\ \#A\# Plan\\ - Input: the problem given\\ - Functionality: give a plan on how to solve the question, including a sketch of the solution, libs to be used,\\ and code snippets\\ - Output: text description of the plan and potential code snippets of the form\\     ```python\\     \textless{}your code\textgreater\\     ```\\ \\ \#A\# Build FOL model\\ - Input: the original problem given\\ - Functionality: build the FOL system that represents the problem; use check\_constraint or check\_model to produce\\   output\\ - Output: the main() function with z3 code of the FOL system of the form\\     ```python\\     def main():\\         \textless{}your code\textgreater\\     ```\\ \\ \#A\# Revise code\\ - Input: z3 code built so far, with potential feedbacks from observations or users\\ - Functionality: reflect on the Input, specify if the tactic is good so far, and if not what are the issues;\\   then, revise the code to continue the problem-solving process or address the issues.\\ - Output: the main() function with revised z3 code of the form\\     ```python\\     def main():\\         \textless{}your code\textgreater\\     ```\\ \\ \#A\# Aggregate and answer\\ - Input: all z3 code, revisions, and observations so far\\ - Functionality: aggregate and summarize the outputs produced so far, and provide the problem-specific final answer\qquad\qquad\qquad\\ - Output: the problem-specific answer\\ \\ \#A\# Tactic check\\ - Input: the original problem, all z3 code, revisions, and observations so far\\ - Functionality: analyze the Input, determine if the tactic can solve the problem or not\\ - Output: "Tactic Good" if tactic can solve the problem; "Tactic Bad" if tactic cannot solve the\\   problem. \\ \bottomrule
\end{xtabular}%
\captionof{table}{Tactic description of predicate logic.}
\vspace{20mm}
\label{tab:tactic-doc}
}

{\small\tabcolsep=3pt  
\centering
\begin{xtabular}{@{}l@{}}
\toprule \\
=== Question and answer\\ \\ Answer the question below by choosing the correct statement.\\ \\ \#\#\# Context:\\ In a study, pairs of trained dogs were placed side by side and given a command such as "sit. " After both obeyed the command, \\ one dog was given a treat while its partner was given no reward at all. Over time, the dogs who went unrewarded began to disobey\\  the command. This shows that dogs have an aversion to being treated unfairly.\\ \\ \#\#\# Question:\\ Which one of the following would be most useful to know in order to evaluate the argument?\\ \\ \#\#\# Statements:\\ 1. Were dogs who were accustomed to receiving regular rewards prior to the study more inclined to obey the command?\\ 2. How many repetitions were required before the unrewarded dogs began to disobey the command?\\ 3. Is there a decline in obedience if rewards are withheld from both dogs in the pair?\\ 4. Were dogs who received treats in one trial ever used as dogs that did not receive treats in other trials?\\ \\ 
Answer: 3\\ \\ === Proposed program\\ \\ ```python\\ class DogBehavior:\\     def \_\_init\_\_(self, accustomed\_reward, repetitions\_until\_disobey, responses\_without\_rewards, consistency\_of\_treatment):\\         self.accustomed\_reward = accustomed\_reward\\         self.repetitions\_until\_disobey = repetitions\_until\_disobey\\         self.responses\_without\_rewards = responses\_without\_rewards\\         self.consistency\_of\_treatment = consistency\_of\_treatment\\     def evaluate\_statement\_based\_on\_behavior(self, statement\_number):\\         if statement\_number == 1:\\             \qquad return self.accustomed\_reward\\         elif statement\_number == 2:\\            \qquad  return self.repetitions\_until\_disobey\\         elif statement\_number == 3:\\             \qquad return self.responses\_without\_rewards\\         elif statement\_number == 4:\\             \qquad return self.consistency\_of\_treatment\\ accustomed\_reward = True \\ repetitions\_until\_disobey = 10 \\ responses\_without\_rewards = False \\ consistency\_of\_treatment = False \\ dog\_behavior = DogBehavior(accustomed\_reward, 
repetitions\_until\_disobey, responses\_without\_rewards, consistency\_of\_treatment)\\ outcome\_1 = dog\_behavior.evaluate\_statement\_based\_on\_behavior(1)\\ outcome\_2 = dog\_behavior.evaluate\_statement\_based\_on\_behavior(2)\\ outcome\_3 = dog\_behavior.evaluate\_statement\_based\_on\_behavior(3)\\ outcome\_4 = dog\_behavior.evaluate\_statement\_based\_on\_behavior(4)\\ print(f"Evaluation of statement 1 leads to: \{outcome\_1\}")\\ print(f"Evaluation of statement 2 leads to: \{outcome\_2\}")\\ print(f"Evaluation of statement 3 leads to: \{outcome\_3\}")\\ print(f"Evaluation of statement 4 leads to: \{outcome\_4\}")\\ ```\\ This code doesn't actually solve the problem and is intended for illustration only. It is meant to proceed with the tactic and show how \\ we could evaluate the significance of statements if we had a more concrete model and data.\\ \\ === outputs\\ \\ \# Observer: python interpreter\\ \# Feedback status: feedback ok\\ \# Content:\\ stdout:\\ Evaluation of statement 1 leads to: True\\ Evaluation of statement 2 leads to: 10\\ Evaluation of statement 3 leads to: False\\ Evaluation of statement 4 leads to: False\\ \\ \#\#\# Comments\\ This is a trivial program that assigns answers to the statements without representing them in details\\ \#\#\# Program good\\ N
\\ \midrule

=== Question and answer\\ \\ Answer the question below by choosing the correct statement.\\ \\ \#\#\# Context:\\ Light is registered in the retina when photons hit molecules of the pigment rhodopsin and change the molecules' shape. Even when they have \\ not been struck by photons of light, rhodopsin molecules sometimes change shape because of normal molecular motion, thereby \\ introducing error into the visual system. The amount of this molecular motion is directly proportional to the temperature of the retina.\\ \\ \#\#\# Question:\\ Which one of the following conclusions is most strongly supported by the information above?\\ \\ \#\#\# Statements:\\ 1. Molecules of rhodopsin are the only pigment molecules that occur naturally in the retina.\\ 2. The visual systems of animals whose body temperature matches that of their surroundings are more error-prone in hot surroundings than \\in cold ones.\\ 3. As the temperature of the retina rises, rhodopsin molecules react more slowly to being struck by photons.\\ 4. The temperature of an animal's retina depends on the amount of light the retina is absorbing.\\ \\ Answer: 2\\ \\ === Proposed program\\ \\ 
```python\\ 
import numpy as np\\
def molecular\_motion(temperature):\\ 
\qquad error\_rate = np.poly1d({[}1, 0{]})(temperature)\\ 
\qquad return error\_rate\\ temperatures = np.array(range(0, 100))\\ error\_rates = molecular\_motion(temperatures)\\ consistent\_statement\_1 = False\\ consistent\_statement\_2 = True\\ consistent\_statement\_3 = False\\ consistent\_statement\_4 = False\\ print(f"Consistent with Statement 2: \{consistent\_statement\_2\}")\qquad\qquad\qquad\qquad\qquad\qquad\qquad\qquad\qquad\qquad\qquad\qquad\qquad\qquad\qquad\qquad\\ 
```\\ 
\\ === outputs\\ \\ \# Observer: python interpreter\\ \# Feedback status: feedback ok\\ \# Content:\\ stdout:\\ Consistent with Statement 2: True\\ \\ \#\#\# Comments\\ This program trivially represents all the statements\\ \#\#\# Program good\\ N \\ \bottomrule
\end{xtabular}%
\captionof{table}{Examples of trivial programs.}
\label{tab:trivial-prog}
\vspace{20mm}
}

{\small\tabcolsep=3pt  
\centering
\begin{xtabular}{@{}l@{}}
\toprule
Hybrid Problem Examples
                                                           \\ \midrule

Answer the question below.\\ \\ \#\#\# Question:\\ Becky has 50 necklaces in her jewelry collection. 3 of the necklaces have broken beads so she collects the unbroken beads for crafting and \\ throws the other parts of the 3 the necklaces out. Becky buys 5 new necklaces that week. She decides to give 15 of her old necklaces to \\ her friends as gifts. How many necklaces does she own now?\\ 37\\ ---\\ Answer the question below.\\ \\ \#\#\# Question:\\ There were 15 males and 10 more girls at the party. Each attendee received 2 cans of soft drinks. If Mary bought several boxes of soft \\ drinks where a box contains 8 cans and is priced at \$5 for each box, how much did Mary spend on soft drinks?\\ 50\\ \\ ===\\ qtype: gg, shuffle: False, label: 1 \\ \\ Becky's jewelry collection sees a variety of activity; she starts with 50 necklaces. Amid handling 3 broken necklaces, she salvages the \\ unbroken beads for crafting purposes. Furthermore, she expands her collection by purchasing 5 new necklaces while generously \\ giving 15 old necklaces to friends, continually rejuvenating her collection’s character.\\ \\ At the vibrant social gathering, there were 15 males and 10 more girls, totaling an attendance of 25 individuals. Each person at the party \\ was provided with 2 cans of soft drinks, ensuring everyone could enjoy their time refreshingly.\\ \\ Y  1. If a box contains 8 cans and is priced at \$5, Mary spent \$50 on soft drinks.\\ N  2. Becky now owns 35 necklaces.

\\ \midrule

Answer the question below.\\ \\ \#\#\# Question:\\ Mark is 18 years old. He has a little brother, John, who is 10 years younger. If John and Mark's parents are \qquad\qquad\qquad\qquad\qquad\\ currently 5 times older than John, how old were they when Mark was born?\\ \\ 22\\ ---\\ Given a set of premises and a hypothesis, answer if the hypothesis\\ agrees with the premises {[}Agree{]},\\ contradicts with the premises {[}Contradict{]},\\ or neutral with respect to the premises {[}Uncertain{]}.\\ \\ \#\#\# Premises:\\ 1. Jason Kramer is an American music supervisor.\\ 2. Some American radio personalities are also music supervisors. \\ 3. Anyone who hosts a show on a public radio station is a radio personality.\\ 4. Joe Rogan is a radio personality.\\ 5. Jason Kramer hosted a show on a public radio station.\\ \\ \#\#\# Hypotheses:\\ Joe Rogan is American.\\ \\ Uncertain\\ ---\\ Answer the question below by choosing the correct statement.\\ \\ \#\#\# Context:\\ Editorial: The threat of harsh punishment for a transgression usually decreases one' s tendency to feel guilt \\ or shame for committing that transgression, and the tendency to feel guilt or shame for committing a \\ transgression reduces a person' s tendency to commit transgressions. Thus, increasing the severity of the\\  legal penalties for transgressions may amplify people' s tendency to ignore the welfare of others.\\ \\ \#\#\# Question:\\ Which one of the following is an assumption required by the editorial's argument?\\ \\ \#\#\# Statements:\\ 1. The threat of harsh punishment deters people from committing transgressions only if this threat is at least \\ sometimes carried out.\\ 2. Everyone has at least some tendency to feel guilt or shame for committing extremely severe transgressions.\\ 3. People who are concerned about threats to their own well-being tend to be less concerned about the \\ welfare of others.\\ 4. At least some actions that involve ignoring the welfare of others are transgressions.\\ \\ 4\\ \\ ===\\ qtype: gfr, shuffle: True, label: 1 \\ \\ Jason Kramer, an American music supervisor, shares a fascinating linkage with Joe Rogan, both existing in the\\  dynamic sphere of radio personalities. While Joe Rogan is widely recognized for his engaging shows, Kramer \\ is known for his music supervision but also hosted a show on a public radio station, making him a radio \\ personality by definition. This interconnection reflects on others in the industry, as it's noted that some American \\ radio personalities double up as music supervisors, enriching their careers with versatility.\\ \\ In a different setting, Mark, now 18 years old, grows through life’s stages, standing ten years above his younger\\  brother, John. Their family dynamics are captivating as their parents are currently aged at a remarkable five \\ times older than John, placing the parents in a stage of life filled with experience and wisdom. \\ \\ Amid these personal stories, an intriguing discussion surfaces regarding the justice system's approach to \\ transgressions. An editorial suggests that the threat of harsh punishment, rather than cultivating a sense of remorse \\ or deterring wrongdoing, might actually reduce feelings of guilt or shame associated with transgressions. \\ Consequently, it posits that escalating legal penalties could unintentionally promote neglect for the welfare of \\ others, pointing to a complex interplay between law, emotion, and societal behavior. This narrative weaves \\ through the lives of individuals, questioning how societal structures influence personal and professional lives.\\ \\ Y 1. "Joe Rogan is American." is Uncertain to the passage above\\ N 2. When Mark was born, John and Mark's parents were 21 years old.\\ N 3. The assumption "The threat of harsh punishment deters people from committing transgressions only if\qquad\qquad\qquad\qquad\qquad\\  this threat is at least sometimes carried out." is required by the editorial's argument.
 
 \\ \bottomrule
\end{xtabular}%
\captionof{table}{Example hybrid problems. The first problem is blended without shuffling and interleaving, and the second is blended with shuffling and interleaving.}
\label{tab:hyb-problems}
\vspace{20mm}
}

{\small\tabcolsep=3pt  
\centering
\begin{xtabular}{@{}l@{}}
\toprule
Routing trajectory Examples
                                                           \\ \midrule

below are example questions and the steps that solve them\\ \\ === Example question\\ \\ Read the context and choose the correct statement.\\ \\ \#\#\# Context:\\ Fluoride's journey into groundwater commences when rain interacts with soil, dissolving minerals rich in fluoride. \\ Amidst this scientific realm, a fascinating study revealed that when variables such as rainfall and mineral concentrations \\ are stable, areas with high sodium levels in the groundwater portrayed significantly increased fluoride concentrations. \\ This distinct geological scenario interlaces with an academic setting where logical structures hold sway. \\ \\ If someone secures a job at a school, they find themselves on the payroll, a fundamental link established within educational \\ employment regulations. Building on this, all faculty members definitely have a job at a school, thus securing their position \\ on the payroll. It's from here that the dual possibilities for Nancy emerge: if she's a teacher, naturally, she is paid by the school; \\ if not, she remains unpaid. This dichotomy resonates with the vehicular debates in sports cars, known for their loud presence – \\ every sports car defies the silence.\\ \\ Switching focus to specific automotive brands, all cars fashioned in Maranello are indeed Ferraris, embedding a mark of luxury\\  and speed wherein being a Ferrari defaults to being a loud sports car, contrasting sharply with the Prius which is neither a \\ sports car nor inherently loud, suggesting that it might be a Maranello-made model or perhaps, ironically, a loud car under \\ different conditions. \\ \\ Meanwhile, a high school bowling team, engaging in a sport of precision and teamwork, accumulated a total of 810 points\\  in their first competitive outing. The scoring dynamics were intriguing: the first bowler earned 1/3 the points of the second,\\  who in turn scored three times higher than the third. In these educational and recreational settings, logical sequences and \\ dependencies manifest vividly, drawing parallels to the environmental phenomenon of fluoride transmission through nature's \\ unassuming actions.\\ \\ \#\#\# Statements:\\ 1. "Nancy is paid by the school and has students." Contradicts the passage above\\ 2. "If Prius is a Ferrari or a loud car, then Prius is an electric car." Contradicts the passage above\\ 3. The third bowler scored 159 points.\\ 4. The statement "Fluoride-bearing minerals are not the primary source of fluoride found in groundwater." can most \\ reasonably be concluded on the basis of the researchers' findings.\\ \\ === Final answer is one of the following\\ \\ 1, 2, 3, 4\\ \\ === Example steps\\ \\ \#\#\# Thought\\ Option 1 is a logic problem. I will use formal logic tactic to solve it.\qquad\qquad\qquad\qquad\qquad\qquad\qquad\qquad\qquad\qquad\\ \#\#\# Action\\ \#\# Name\\ Call tactic: formal logic z3\\ \#\# Input\\ The main problem, and the past results\\ \#\# Output\\ \#\#\# option\\ 1\\ \#\#\# subproblem\\ Given a set of premises and a hypothesis, answer if the hypothesis\\ agrees with the premises {[}Agree{]},\\ contradicts with the premises {[}Contradict{]},\\ or neutral with respect to the premises {[}Uncertain{]}.\\ \\ \#\#\# Premises:\\ 1. If you are on the payroll, then you are being paid by the school.\\ 2. If someone has a job at a school, then they are on the payroll.\\ 3. All faculty members have a job at a school.\\ 4. One can either be a faculty member or a teacher.\\ 5. Every teacher has students.\\ 6. If Nancy is a teacher, then they are on the payroll.\\ 7. If Nancy is not a teacher, then they are not paid by the school.\\ \\ \#\#\# Hypotheses:\\ Nancy is paid by the school and has students.\\ \\ === observations ===\\ \\ \# Observer: Action parser\\ \# Feedback status: feedback ok\\ \# Content:\\ Solving subproblem with tactic predicate\_logic\_z3\\ \\ \# Observer: Runner\\ \# Feedback status: feedback ok\\ \# Content:\\ Tactic execution successful. Tactic output:\\ Agree\\ \\ === response ===\\ \\ \#\#\# Thought\\ Option 2 is a logic problem. I will use formal logic tactic to solve it.\\ \#\#\# Action\\ \#\# Name\\ Call tactic: formal logic z3\\ \#\# Input\\ The main problem, and the past results\\ \#\# Output\\ \#\#\# option\\ 2\\ \#\#\# subproblem\\ Given a set of premises and a hypothesis, answer if the hypothesis\\ agrees with the premises {[}Agree{]},\\ contradicts with the premises {[}Contradict{]},\\ or neutral with respect to the premises {[}Uncertain{]}.\\ \\ \#\#\# Premises:\\ 1. All sports cars are loud.\\ 2. No loud cars are electric.\\ 3. If a car is a Ferrari, then it is a sports car.\\ 4. All cars made in Maranello are Ferrari\\ 5. If Prius is neither a sports car nor a loud car, then Prius is a Maranello-made car or a loud car.\\ \\ \#\#\# Hypotheses:\\ If Prius is a Ferrari or a loud car, then Prius is an electric car.\\ \\ === observations ===\\ \\ \# Observer: Action parser\\ \# Feedback status: feedback ok\\ \# Content:\\ Solving subproblem with tactic predicate\_logic\_z3\\ \\ \# Observer: Runner\\ \# Feedback status: feedback ok\\ \# Content:\\ Tactic execution successful. Tactic output:\\ Contradict\\ \\ === response ===\\ \\ \#\#\# Thought\\ Option 3 is a math problem. I will use math tactic to solve it.\\ \#\#\# Action\\ \#\# Name\\ Call tactic: math\\ \#\# Input\\ The main problem, and the past results\\ \#\# Output\\ \#\#\# option\\ 3\\ \#\#\# subproblem\\ Answer the question below.\\ \\ \#\#\# Question:\\ A high school bowling team's 3 members scored a total of 810 points in their first match.  The first bowler scored \\ 1/3 as many points as the second, and the second bowler scored 3 times as high as the third bowler.  How many \\ points did the third bowler score?\\ \\ === observations ===\\ \\ \# Observer: Action parser\\ \# Feedback status: feedback ok\\ \# Content:\\ Solving subproblem with tactic math\\ \\ \# Observer: Runner\\ \# Feedback status: feedback ok\\ \# Content:\\ Tactic execution successful. Tactic output:\\ 162\\ \\ === response ===\\ \\ \#\#\# Thought\\ Option 4 is a commonsense reasoning problem. I will use general program tactic to solve it.\\ \#\#\# Action\\ \#\# Name\\ Call tactic: general program\\ \#\# Input\\ The main problem, and the past results\\ \#\# Output\\ \#\#\# option\\ 4\\ \#\#\# subproblem\\ Answer the question below by choosing the correct statement.\\ \\ \#\#\# Context:\\ Delta green ground beetles sometimes remain motionless for hours at a stretch, although they are more active in \\ wet years than in dry years. In 1989 an observer spotted ten delta green ground beetles in nine hours; in 1985 the \\ same observer at the same location had counted 38 in about two hours. This difference probably does not reflect \\ a drop in the population of these rare beetles over this period, however, because 1985 was a wet year and 1989\\  was relatively dry.\\ \\ \#\#\# Question:\\ Which one of the following, if true, most strongly supports the conclusion drawn above?\\ \\ \#\#\# Statements:\\ 1. Fluoride-bearing minerals are not the primary source of fluoride found in groundwater.\\ 2. None of the above\\ \\ === observations ===\\ \\ \# Observer: Action parser\\ \# Feedback status: feedback ok\\ \# Content:\\ Solving subproblem with tactic any\_program\\ \\ \# Observer: Runner\\ \# Feedback status: feedback ok\\ \# Content:\\ Tactic execution successful. Tactic output:\\ 2\\ \\ === response ===\\ \\ \#\#\# Thought\\ I have solved all the subproblems, I will aggregate the results and produce the answer\\ \#\#\# Action\\ \#\# Name\\ Aggregate and answer\\ \#\# Input\\ all thoughts, actions, and observations so far\qquad\qquad\qquad\qquad\qquad\qquad\qquad\qquad\qquad\qquad\qquad\qquad\qquad\qquad\qquad\qquad\\ \#\# Output\\ 2
 
 \\ \bottomrule
\end{xtabular}%
\captionof{table}{An example routing trajectory.}
\label{tab:rout-traj}
}

\end{document}